\documentclass[11pt]{article}
\usepackage[utf8]{inputenc}
\usepackage[english]{babel}
\usepackage{tikz}
\usetikzlibrary{chains, positioning}
\usepackage{enumerate}
\usepackage{color}
\usepackage{graphicx}
\usepackage{amssymb, amsmath, amsthm}
\usepackage{amsfonts}
\usepackage{bm}
\usepackage{calc}
\usepackage{caption}
\usepackage{subcaption}
\usepackage{url}
\usepackage{epstopdf}
\usepackage{float}

\usepackage{geometry}
\usepackage[title]{appendix}
\usepackage{authblk}

\numberwithin{equation}{section}

\theoremstyle{remark}
\newtheorem{remark}{Remark}[section]

\theoremstyle{definition}

\theoremstyle{definition}
\newtheorem*{definition*}{Definition}

\DeclareMathOperator*{\argmin}{arg\,min}

\author{Jens Berg\thanks{jens.berg@math.uu.se}\ }
\author{Kaj Nystr\"{o}m\thanks{kaj.nystrom@math.uu.se}}
\affil{Department of Mathematics, Uppsala University\\
S-751 06 Uppsala, Sweden}

\title{A unified deep artificial neural network approach to partial differential equations in complex geometries}
\date{}

\begin{document}

\maketitle

\begin{abstract}
In this paper we use deep feedforward artificial neural networks to approximate solutions to partial differential equations in complex geometries. We show how to modify the backpropagation algorithm to compute the partial derivatives of the network output with respect to the space variables which is needed to approximate the differential operator.  The method is based on an ansatz for the solution which requires nothing but feedforward neural networks and an unconstrained gradient based optimization method such as gradient descent or a quasi-Newton method.

We show an example where classical mesh based methods cannot be used and neural networks can be seen as an attractive alternative. Finally, we highlight the benefits of deep compared to shallow neural networks and device some other convergence enhancing techniques.
\end{abstract}

 \setcounter{equation}{0} \setcounter{theorem}{0}

\section{Introduction}
Partial differential equations (PDEs) are used to model a variety of phenomena in the natural sciences. Common to most of the PDEs encountered in practical applications is that they cannot be solved analytically but require various approximation techniques. Traditionally, mesh based methods such as finite elements (FEM), finite differences (FDM), or finite volumes (FVM), are the dominant techniques for obtaining approximate solutions. These techniques require that the computational domain of interest is discretized into a set of mesh points and the solution is approximated at the points of the mesh. The advantage of these methods is that they are very efficient for low-dimensional problems on regular geometries. The drawback is that for complicated geometries, meshing can be as difficult as the numerical solution of the PDE itself. Moreover, the solution is only computed at the mesh points and evaluation of the solution at any other point requires interpolation or some other reconstruction method.

In contrast, other methods do not require a mesh but a set of collocation points where the solution is approximated. The collocation points can be generated according to some distribution inside the domain of interest and examples include radial basis functions (RBF) and Monte Carlo methods (MCM). The advantage is that it is relatively easy to generate collocation points inside the domain by a hit-and-miss approach. The drawbacks of these methods compared to traditional mesh based methods are for example numerical stability for RBF and inefficiency for MCM.

The last decade has seen a revolution in deep learning where deep artificial neural networks (ANNs) are the key component. ANNs have been around since the 40's \cite{firstnet} and used for various applications. A historical review, particularly in the context of differential equations, can be found in ch. 2 in \cite{yadav2015introduction}. The success of deep learning during the last decade is due to a combination of improved theory starting with unsupervised pre-training and deep belief nets and improved hardware resources such as general purpose graphics processing units (GPGPUs). See for example \cite{firstdeep, imagenetwin}. Deep ANNs are now routinely used with impressive results in areas such as image analysis, pattern recognition, object detection, natural language processing, and  self-driving cars to name a few areas.

While deep ANNs have achieved impressive results in several important application areas there are still many open questions concerning how and why they actually work so efficiently. From the perspective of function approximation theory it has been known since the 90's that ANNs are universal approximators that can be used to approximate any continuous function and its derivatives \cite{Hornik1989359, Hornik1990551, 80265, Li1996327}. In the context of PDEs single hidden layer ANNs have traditionally been used to solve PDEs since one hidden layer with sufficiently many neurons is sufficient for approximating any function and as all gradients that are needed can be computed in analytical closed form \cite{lagarisold, lagarisnew, 5061501}. More recently there is a limited but emerging literature on the use of deep ANNs to solve PDEs \cite{Rudd2014571, Rudd2015277, deephigh, deeprelaxation, 2017arXiv171110561R, 2017arXiv171000211E, dgm}. In general ANNs have the benefits that they are smooth, analytical functions which can be evaluated at any point inside, or even outside, the domain without reconstruction.

In this paper we introduce a method of solving PDEs which only involves feedforward deep ANNs with (close to) no user intervention. In previous works only a single  hidden layer is used and the method outlined requires some user-defined functions which are non-trivial or even impossible to compute. See Lagaris et. al \cite{lagarisold}. Later Lagaris et. al \cite{lagarisnew} removed the need for user intervention by using a combination of feedforward and radial basis function ANNs in a non-trivial way. Later work by McFall and Mahan \cite{5061501} removed the need for the radial basis function ANN by replacing it with length factors that are computed using thin plate splines. The thin plate splines computation is, however, rather cumbersome as it involves the solution of many linear systems which adds extra complexity to the problem. In contrast, the method proposed in this paper requires only an implementation of a deep feedforward ANN together with a cost function and an arbitrary choice of unconstrained numerical optimization method.

 We denote the method presented in this paper as \textit{unified} as it only requires feedforward neural networks in contrast to the previous methods presented by Lagaris et. al and McFall et. al. We are not claiming that ANNs necessarily are suitable for solving PDEs in low-dimensions and simple geometries where they are outperformed by classical mesh based methods. Instead, the method have its merits for high-dimensional problems and complex domains where most numerical PDE techniques become infeasible.

As training neural networks consumes a lot of time and computational resources it is desirable to reduce the number of required iterations until a certain accuracy has been reached. During this work we have found two factors which strongly influence the number of required iterations in our PDE focused applications/examples. The first is pre-training of the network using the available boundary data and the second is to increase the number of hidden layers. One of our particular findings is that we, by fixing the capacity of the network and increasing the number of hidden layers, can see a dramatic decrease in the number of iterations required to reach a desired level of accuracy. Our findings indicate that the use of deep ANNs instead of just shallow ones adds real value in the context of using ANNs to solve PDEs.

 The rest of the paper is organized as follows. In section \ref{sec2} we recall the network architecture of deep fully connected feedforward ANNs, introduce some necessary notation and the backpropagation scheme. In section \ref{sec3} we develop unified artificial neural network approximations for stationary PDEs. Based on our ansatz for the solution we discuss extension of the boundary data, smooth distance function approximation and computation, gradient computations and the backpropagation algorithm for parameter calibration. The detailed account of the latter computations/algorithms in the case of advection and diffusion problems are given in Appendix A and B, respectively. In section \ref{sec4} we provide some concrete examples concerning how to apply the method outlined. Our examples include linear advection and diffusion in 1D and 2D, but high-dimensional problems are also discussed. Section \ref{sec5} is devoted to the convergence considerations briefly mentioned above and we here highlight that the use of deep ANNs instead of just shallow seems to add real value in the context of using ANNs to solve PDEs. Finally, section \ref{sec7} is devoted to a summary and conclusions.

\section{Network architecture}\label{sec2}
In this paper we consider deep, fully connected feedforward ANNs. The ANN consists of $L + 1$ layers where layer 0 is the input layer and layer $L$ is the output layer. The layers $0 < l < L$ are the hidden layers. The activation functions in the hidden layers can be any activation function such as for example sigmoids, rectified linear units, or hyperbolic tangents. Unless otherwise stated we will use sigmoids in the hidden layers. The output activation will be the linear activation. The ANN defines a mapping $\mathbb{R}^N \to \mathbb{R}^M$.

Each neuron in the ANN is supplied with a bias, including the output neurons but excluding the input neurons, and the connections between neurons in subsequent layers are represented by matrices of weights. We let $b^l_j$ denote the bias of neuron $j$ in layer $l$. The weight between neuron $k$ in layer $l-1$ and neuron $j$ in layer $l$ is denoted by $w^l_{jk}$. The activation function in layer $l$ will be denoted by $\sigma_l$ regardless of the type of activation. We assume for simplicity that a single activation function is used for each layer. The output from neuron $j$ in layer $l$ will be denoted by $y^l_j$. See Figure~\ref{network} for a schematic representation of a fully connected feedforward ANN.

\tikzset{
  every neuron/.style={
    circle,
    draw,
    minimum size=1.2cm
  },
  neuron missing/.style={
    draw=none,
    scale=2,
    text height=0.25cm,
    execute at begin node=\color{black}$\vdots$
  },
}

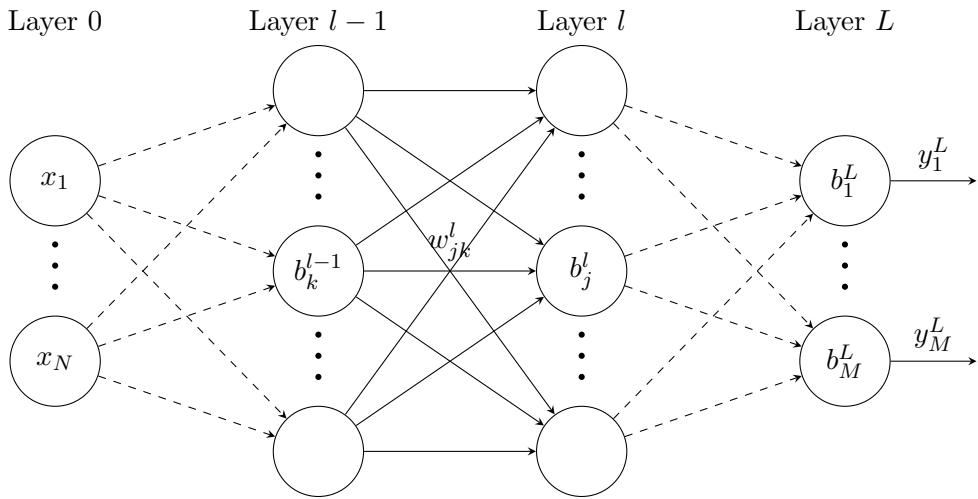
\begin{figure}
\centering
\begin{tikzpicture}[x=1.75cm, y=1.2cm, >=stealth]

\foreach \m/\l [count=\y] in {1,missing,2}
	\node [every neuron/.try, neuron \m/.try] (input-\m) at (0,1.5-\y) {};

\foreach \m [count=\y] in {1,missing}
	\node [every neuron/.try, neuron \m/.try ] (hidden1-\m) at (2,2.5-\y) {};
\foreach \m [count=\y] in {2}
	\node [every neuron/.try, neuron \m/.try ] (hidden1-\m) at (2,2.5-\y-2) {$b^{l-1}_k$};
\foreach \m [count=\y] in {missing,3}
	\node [every neuron/.try, neuron \m/.try ] (hidden1-\m) at (2,2.5-\y-3) {};

\foreach \m [count=\y] in {1,missing}
	\node [every neuron/.try, neuron \m/.try ] (hidden2-\m) at (4,2.5-\y) {};
\foreach \m [count=\y] in {2}
	\node [every neuron/.try, neuron \m/.try ] (hidden2-\m) at (4,2.5-\y-2) {$b^l_j$};
\foreach \m [count=\y] in {missing,3}
	\node [every neuron/.try, neuron \m/.try ] (hidden2-\m) at (4,2.5-\y-3) {};

\foreach \m [count=\y] in {1,missing,2}
  \node [every neuron/.try, neuron \m/.try ] (output-\m) at (6,1.5-\y) {};

\foreach \l [count=\i] in {1,N}
    \node [] at (input-\i) {$x_{\l}$};

\node [] at (output-1) {$b^L_1$};
\node [] at (output-2) {$b^L_M$};

\foreach \l [count=\i] in {1,M}
  \draw [->] (output-\i) -- ++(1,0)
    node [above,midway] {$y^L_\l$};

\foreach \i in {1,...,2}
  \foreach \j in {1,...,3}
    \draw [dashed,->] (input-\i) -- (hidden1-\j);

\foreach \i in {1}
  \foreach \j in {1,...,3}
    \draw [->] (hidden1-\i) -- (hidden2-\j);

\foreach \i in {2}
  \foreach \j in {1}
    \draw [->] (hidden1-\i) -- (hidden2-\j);

\foreach \i in {2}
  \foreach \j in {2}
    \draw [->] (hidden1-\i) -- (hidden2-\j)
    	node [above,midway] {$w^l_{jk}$};

\foreach \i in {2}
  \foreach \j in {3}
    \draw [->] (hidden1-\i) -- (hidden2-\j);

\foreach \i in {3}
  \foreach \j in {1,...,3}
    \draw [->] (hidden1-\i) -- (hidden2-\j);

\foreach \i in {1,...,3}
  \foreach \j in {1,...,2}
    \draw [dashed,->] (hidden2-\i) -- (output-\j);

\foreach \l [count=\x from 0] in {Layer 0, Layer $l-1$, Layer $l$, Layer $L$}
  \node [align=center, above] at (\x*2,2) {\l};

\end{tikzpicture}
\caption{Schematic representation of a fully connected feedforward ANN.}
\label{network}
\end{figure}

A quantity that will be used extensively is the so-called weighted input which is defined as
\begin{equation}
z^l_j = \sum_k w^l_{jk}\sigma_{l-1}(z^{l-1}_k) + b^l_j
\label{wi}
\end{equation}
where the sum is taken over all inputs to neuron $j$ in layer $l$. That is, the number of neurons in layer $l-1$. The weighted input (\ref{wi}) can of course also be written in terms of the output from the previous layer as
\begin{equation}
z^l_j = \sum_k w^l_{jk}y^{l-1}_k + b^l_j
\end{equation}
where the output $y^{l-1}_k = \sigma_{l-1}(z^{l-1}_k)$ is the activation of the weighted input.
As we will be working with deep ANNs we will prefer formula (\ref{wi}) as it naturally defines a recursion in terms of previous weighted inputs through the ANN. By definition we have
\begin{equation}
\sigma_0(z^0_j) = y^0_j = x_j
\end{equation}
which terminates any recursion.

By dropping the subscripts we can write (\ref{wi}) in the convenient vectorial form
\begin{equation}
z^l = W^l\sigma_{l-1}(z^{l-1}) + b^l= W^ly^{l-1} + b^l
\end{equation}
where each element in the $z^l$ and $y^l$ vectors are given by $z^l_j$ and $y^l_j$, respectively, and the activation function is applied elementwise. The elements of the matrix $W^l$ are given by $W^l_{jk} = w^l_{jk}$.

With the above definitions the feedforward algorithm for computing the output $y^L$, given the input $x$, is given by
\begin{equation}
\begin{aligned}
y^L &= \sigma_L(z^L) \\
z^L &= W^L\sigma_{L-1}(z^{L-1}) + b^L \\
z^{L-1} &= W^{L-1}\sigma_{L-2}(z^{L-2}) + b^{L-1} \\
&\mathrel{\makebox[\widthof{=}]{\vdots}} \\
z^{2} &= W^2\sigma_1(z^1) + b^2 \\
z^{1} &= W^1x + b^1.
\end{aligned}
\label{feedforward}
\end{equation}

\subsection{Backpropagation}
Given some data $x = [x_1, \ldots, x_N]^T \in \mathbb{R}^N$ and some target outputs $y = [y_1, \ldots, y_M]^T \in \mathbb{R}^M$ we wish to choose our weights and biases such that $y^L(x; w, b)$ is a good approximation of $y(x)$. We use the notation $y^L(x; w, b)$ to indicate that the ANN takes $x$ as input and is parametrized by the weights and biases $w$, $b$. To find the weights and biases we define some cost function $C = C(y, y^L): \mathbb{R}^N \to \mathbb{R}$ and compute
\begin{equation}
w^*, b^* = \argmin_{w, b} C(y, y^L).
\label{minprop}
\end{equation}
The minimization problem (\ref{minprop}) can be solved using a variety of optimization methods, both gradient based and gradient free. In this paper we will focus on gradient based methods and we derive the gradients that we need in order to use any gradient based optimization method. The standard method to compute gradients is backpropagation. The main ingredient is the error of neuron $j$ in layer $l$ defined by
\begin{equation}
\delta^l_j = \frac{\partial C}{\partial z^l_j}.
\end{equation}
The standard backpropagation algorithm for computing the gradients of the cost function is then given by
\begin{equation}
\begin{aligned}
\delta^L_j &= \frac{\partial C}{\partial y^L_j} \sigma'_L(z^L_j), &
\frac{\partial C}{\partial w^l_{jk}} &= y^{l-1}_k \delta^l_j, \\
\delta^l_j &= \sum_k w^{l+1}_{kj} \delta^{l+1}_k \sigma'_l(z^l_j), &
\frac{\partial C}{\partial b^l_j} &= \delta^l_j.
\end{aligned}
\label{backprop}
\end{equation}
The $\delta$-terms in (\ref{backprop}) can be written in vectorial form as
\begin{equation}
\begin{aligned}
\delta^L &= \nabla_{y^L} C \odot \sigma'_L(z^L), &
\delta^l &= (W^{l+1})^T \delta^{l+1} \odot \sigma'_l(z^l)
\end{aligned}
\end{equation}
where we use the notation $\nabla_{y^L} C = [\frac{\partial C}{\partial y^L_1}, \ldots, \frac{\partial C}{\partial y^L_M}]^T$ and $\odot$ denotes the Hadamard (componentwise) product. The notation $\nabla C$ without a subscript will denote the vector of partial derivatives with respect to the input $x = [x_1, \ldots, x_N]^T$. Note the transpose on the weight matrix $W$. The transpose $W^T$ is the Jacobian matrix of the coordinate transformation between layer $l+1$ and $l$.

\section{Unified ANN approximations to PDEs}\label{sec3}
In this section we are interested in solving stationary PDEs of the form
\begin{equation}
\begin{aligned}
Lu &= f, & x &\in \Omega, \\
Bu &= g, & x &\in \Gamma \subset \partial \Omega
\end{aligned}
\label{pde}
\end{equation}
where $L$ is a differential operator, $f$ a forcing function, $B$ a boundary operator, and $g$ the boundary data. The domain of interest is $\Omega \subset \mathbb{R}^N$, $\partial \Omega$ denotes its boundary, and $\Gamma$ is the part of the boundary where boundary conditions should be imposed. In this paper we focus on problems where $L$ is either the advection or diffusion operator, or some mix thereof. The extension to higher order problems are conceptually straightforward based on the derivations outlined below.

We consider the ansatz $\hat{u} = \hat{u}(x; w, b)$ for $u$ where $(w,b)$ denotes the parameters of the underlying ANN. To determine the parameters we will use the cost function defined as the quadratic residual,
\begin{equation}
C = \frac{1}{2}||L\hat{u} - f||^2:= \frac 1 2\int_{\Omega}|L\hat{u} - f|^2\, dx.
\label{rescost}
\end{equation}
For further reference we need the gradient of (\ref{rescost}) with respect to any network parameter. Let $p$ denote any of $w^l_{jk}$ or $b^l_j$. Then
\begin{equation}
\frac{\partial C}{\partial p} = (L\hat{u} - f) \odot (L\frac{\partial \hat{u}}{\partial p}).
\label{rescostgrad}
\end{equation}

We will use the collocation method \cite{lagarisold} and therefore we discretize $\Omega$ and $\Gamma$ into a sets of collocation points $\Omega_d $ and $\Gamma_d$, with $|\Omega_d| = N_d$ and $|\Gamma_d| = N_b$, respectively. In its discrete form the minimization problem (\ref{minprop}) then becomes
\begin{equation}
w^*, b^* = \argmin_{w, b} \sum \limits_{x_i \in \Omega_d} \frac{1}{2}\frac{1}{N_d}||L\hat{u}(x_i) - f(x_i)||^2
\label{collminprob}
\end{equation}
subject to the constraints
\begin{equation}
\begin{aligned}
B\hat{u}(x_i) = g(x_i), && \forall x_i \in \Gamma_d.
\end{aligned}
\label{minconstraints}
\end{equation}

There are many approaches to the minimization problem (\ref{collminprob}) subject to the constraints (\ref{minconstraints}). One can for example use a constrained optimization procedure, Lagrange multipliers, or penalty formulations to include the constraints in the optimization. Another approach is to design the ansatz $\hat{u}$ such that the constraints are automatically fulfilled, thus turning the constrained optimization problem into an unconstrained optimization problem. Unconstrained optimization problems can be more efficiently solved using gradient based optimization techniques.

In the following we let $B$ in (\ref{pde}) be the identity operator and we  hence consider PDEs with Dirichlet boundary conditions. In this case our ansatz for the solution is
\begin{equation}
\hat{u}(x) = G(x) + D(x)y^L(x; w, b)
\label{trialsol}
\end{equation}
where $G=G(x)$ is a smooth extension of the boundary data $g$ and $D=D(x)$ is a smooth distance function giving the distance for $x\in\Omega$ to $\Gamma$.  Note that the form of the ansatz (\ref{trialsol}) ensures that $\hat{u}$ attains its boundary values at the boundary points. Moreover, $G$ and $D$ are pre-computed using low-capacity ANNs for the given boundary data and geometry using only a small subset of the collocation points and do not add any extra complexity when minimizing (\ref{collminprob}). We call this approach unified as there are no other ingredients involved other than feedforward ANNs, and gradient based, unconstrained optimization methods to compute all of $G$, $D$, and $y^L$.

\subsection{Extension of the boundary data}
The ansatz (\ref{trialsol}) requires that $G$ is globally defined, smooth\footnote{It is sufficient that $G$ has continuous derivatives up to the order of the differential operator $L$. However, since $G$ will be computed using an ANN it will be smooth.} and that
\begin{equation}
\begin{aligned}
|G(x) - g(x)| < \epsilon, && \forall x \in \Gamma.
\end{aligned}
\end{equation}
Other than these requirements the exact form of $G$ is not important. It is hence an ideal candidate for an ANN approximation. To compute $G$ we simply train an ANN to fit $g(x)$, $\forall x \in \Gamma_d$. The quadratic cost function used is given by
\begin{equation}
C = \frac{1}{2}\frac{1}{N_b}\sum\limits_{x_i \in \Gamma_d}||G(x_i) - g(x_i)||^2
\end{equation}
and in the course of calibration we compute the gradients using the standard backpropagation (\ref{backprop}). Note that in general we have $N_b << N_d$ and the time it takes to compute $G$ is of lower order.

\subsection{Smooth distance function computation}
To compute the smooth distance function $D$ we start by computing a non-smooth distance function $d$ and approximate it using a low-capacity ANN. For each point $x$ we define $d$ as the minimum distance to a boundary point where a boundary condition should be imposed. That is,
\begin{equation}
d(x) = \min_{x_b \in \Gamma} ||x - x_b||.
\label{distfunc}
\end{equation}
Again, the exact form of $d$ (and $D$) is not important other than that $D$ is smooth and
\begin{equation}
\begin{aligned}
|D(x)| < \epsilon, && \forall x \in \Gamma.
\end{aligned}
\end{equation}
We can use a small subset $\omega_d \subset \Omega_d$ with $|\omega_d| = n_d << N_d$ to compute $d$.

\begin{figure}
\centering
\begin{subfigure}[t]{0.49\textwidth}
\centering
\includegraphics[width=\textwidth]{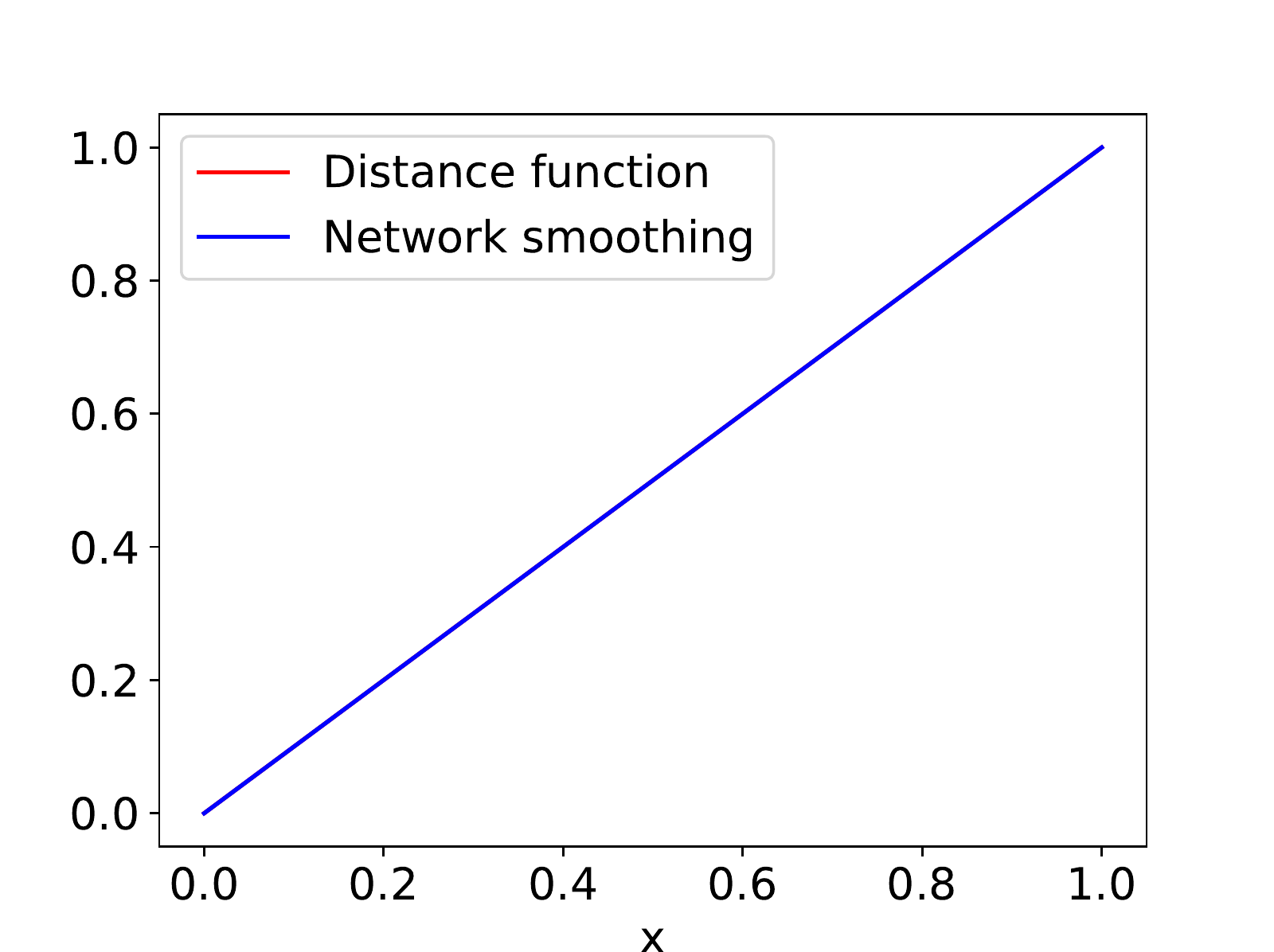}
\caption{$L = d/dx$. Only one boundary condition at $x = 0$.}
\label{distadvec}
\end{subfigure}
\begin{subfigure}[t]{0.49\textwidth}
\centering
\includegraphics[width=\textwidth]{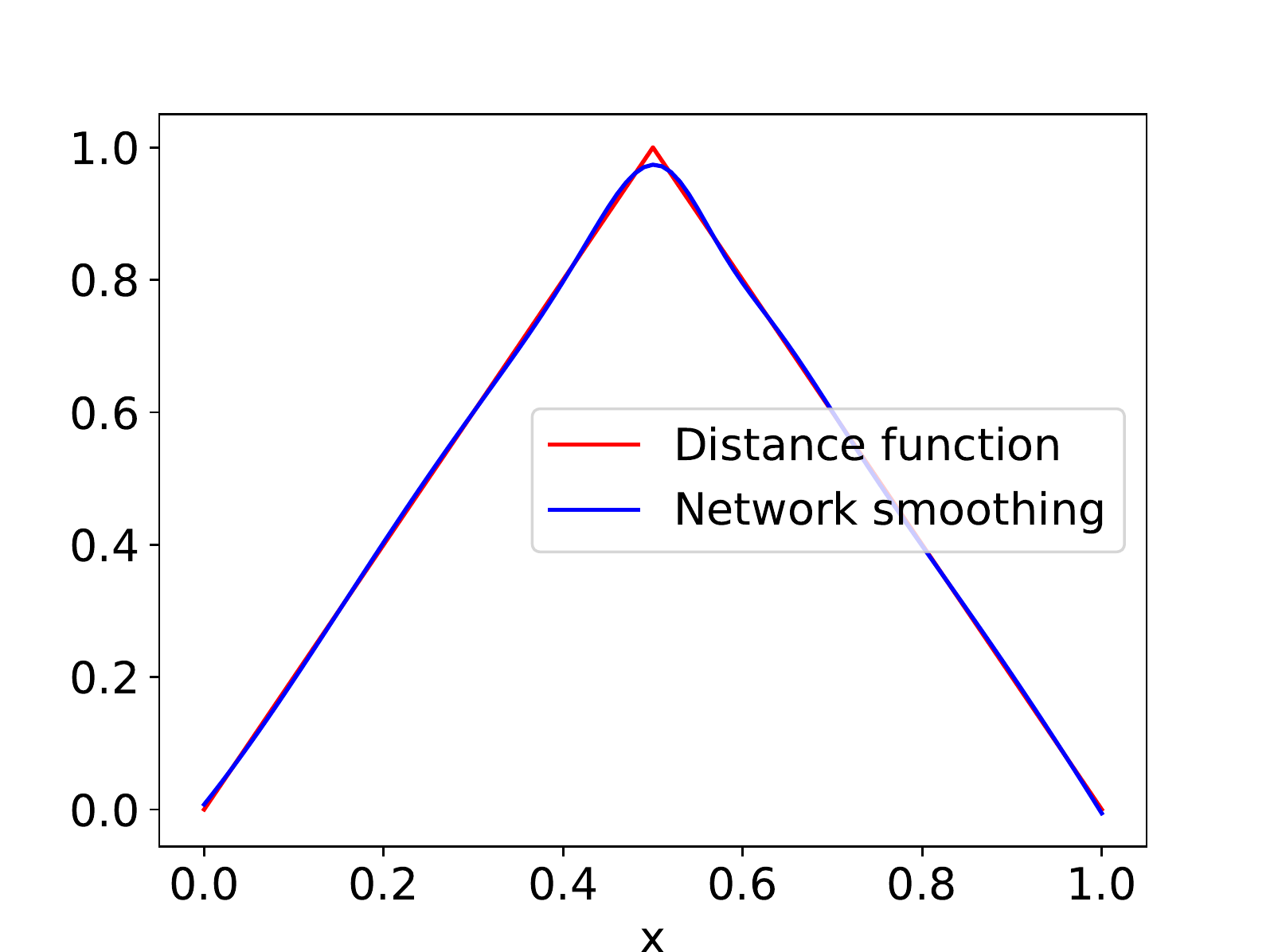}
\caption{$L = d^2/dx^2$. Boundary conditions at both boundary points.}
\label{distheat}
\end{subfigure}
\caption{Smoothed distance functions using single hidden layer ANNs with 5 hidden neurons using 100 collocation points.}
\label{distpic}
\end{figure}

Once $d$ has been computed and normalized we fit another ANN and train it using the cost function
\begin{equation}
C = \frac{1}{2}\frac{1}{n_d + N_b}\sum\limits_{x_i \in \omega_d \cup \Gamma_d}||D(x_i) - d(x_i)||^2.
\end{equation}
Note that we train the ANN over the points $x_i \in \omega_d \cup \Gamma_d$ with $d(x_i) = 0$ $\forall x_i \in \Gamma_d$. Since the ANN defining the smooth distance function can be  taken to be a single hidden layer, low-capacity ANN with $n_d + N_b << N_d$, the total time to compute both $d$ and $D$ is of lower order. See Figure~\ref{distpic} for (rather trivial examples) of $d$ and $D$ in one dimension.

\begin{remark}
We stress that the distance function (\ref{distfunc}) can be computed efficiently with nearest-neighbor searches using for example $k$-d trees \cite{kdtree} or ball trees \cite{balltree} for very high-dimensional input. A na\"{i}ve nearest neighbor search will have $O(N_dN_b)$ complexity while an efficient nearest-neighbor search will typically have $O(N_d \log N_b)$.
\end{remark}

\begin{remark}
Instead of computing the actual distance function we could use the more extreme version
\begin{equation}
d(x) = \begin{cases}
0, & x \in \Gamma, \\
1, & \textnormal{otherwise.}
\end{cases}
\end{equation}
However, simulations showed that this function and its approximating ANN have a negative impact on convergence and quality when computing $\hat{u}$.
\end{remark}

\subsection{Gradient computations}
When the differential operator $L$ acts on the ansatz $\hat{u}$ in (\ref{trialsol}) we need to compute the partial derivatives of the ANNs with respect to the spatial variables $(x_1,...,x_N)$. Also, when using gradient based optimization we need to compute the gradients of the quadratic residual cost function (\ref{rescost}) with respect to the network parameters.

In the single hidden layer case all gradients can be computed in closed analytical form as shown in \cite{lagarisold}. For deep ANNs, however, we need to modify the feedforward (\ref{feedforward}) and backpropagation (\ref{backprop}) algorithms to compute the gradients with respect to $(x_1,...,x_N)$ and network parameters, respectively.

Note that the minimization problem in \eqref{collminprob} involves the collocation points $\{x_i\}$. Throughout the paper we will always denote $\frac{\partial}{\partial x_i}$ as the partial derivative with respect to the spatial coordinate $x_i$, not with respect to collocation point $x_i$.

We supply the details of the relevant gradient computations/algorithms in the case of advection and diffusion problems in Appendix A and B, respectively.

\section{Numerical examples}\label{sec4}
In this section we provide some concrete examples concerning how to apply the modified feedforward and backpropagation algorithms to model problems. Each problem requires their own modified backpropagation algorithms depending on the differential operator $L$. We start by presenting a few simple examples in some detail before we proceed to examples of more complicated problems.

In all of the following numerical examples we have implemented the gradients according to the schemes presented in the previous section and we consequently use the BFGS \cite{bfgs} method with default settings from \texttt{SciPy} \cite{scipy} to train the ANN. We have tried all gradient free and gradient based numerical optimization methods available in the \texttt{scipy.optimize} package, and the online, batch, and stochastic gradient descent methods. BFGS shows superior performance compared to all other methods that we have tried.

An issue one might encounter is that the line search in BFGS fails due to the Hessian matrix being very ill-conditioned. To circumvent this we use a BFGS in combination with stochastic gradient descent. When BFGS fails due to line search failure we run 1000 iterations with stochastic gradient descent with a very small learning rate, in the order of $10^{-9}$, to bring BFGS out of the troublesome region. This procedure is repeated until convergence or until the maximum number of allowed iterations has been exceeded.

The examples in 1D are rather simple and it suffices to use ANNs with two hidden layers with 10 neurons each. The 2D examples are somewhat more complicated and for those we use ANNs with five hidden layers with 10 neurons each.

\subsection{Linear advection in 1D}
\begin{figure}[H]
\centering
\begin{subfigure}[t]{0.49\textwidth}
\includegraphics[width = \textwidth]{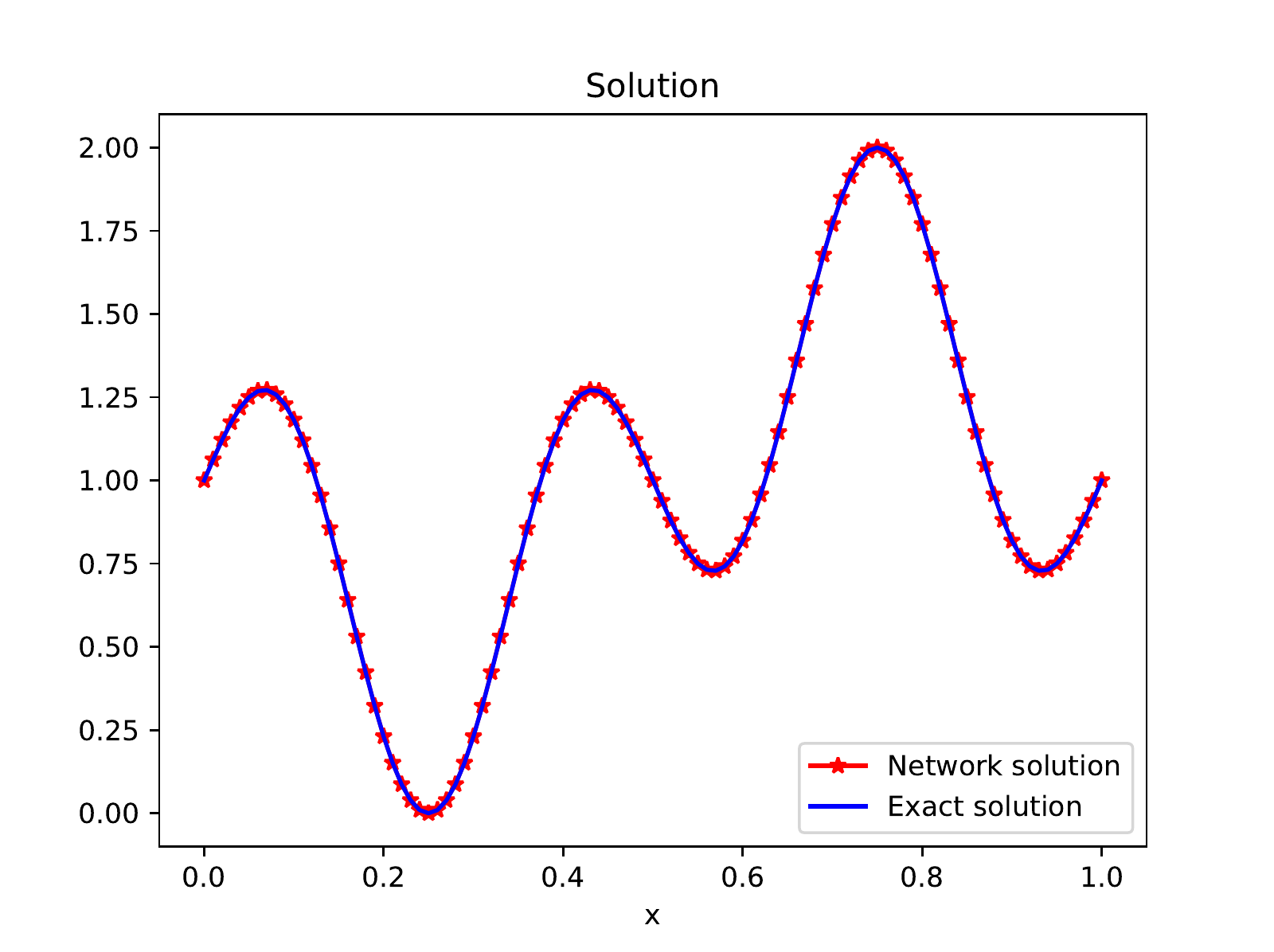}
\caption{Exact and approximate ANN solution to the 1D stationary advection equation.}
\label{advecsol1d}
\end{subfigure}
\begin{subfigure}[t]{0.49\textwidth}
\includegraphics[width = \textwidth]{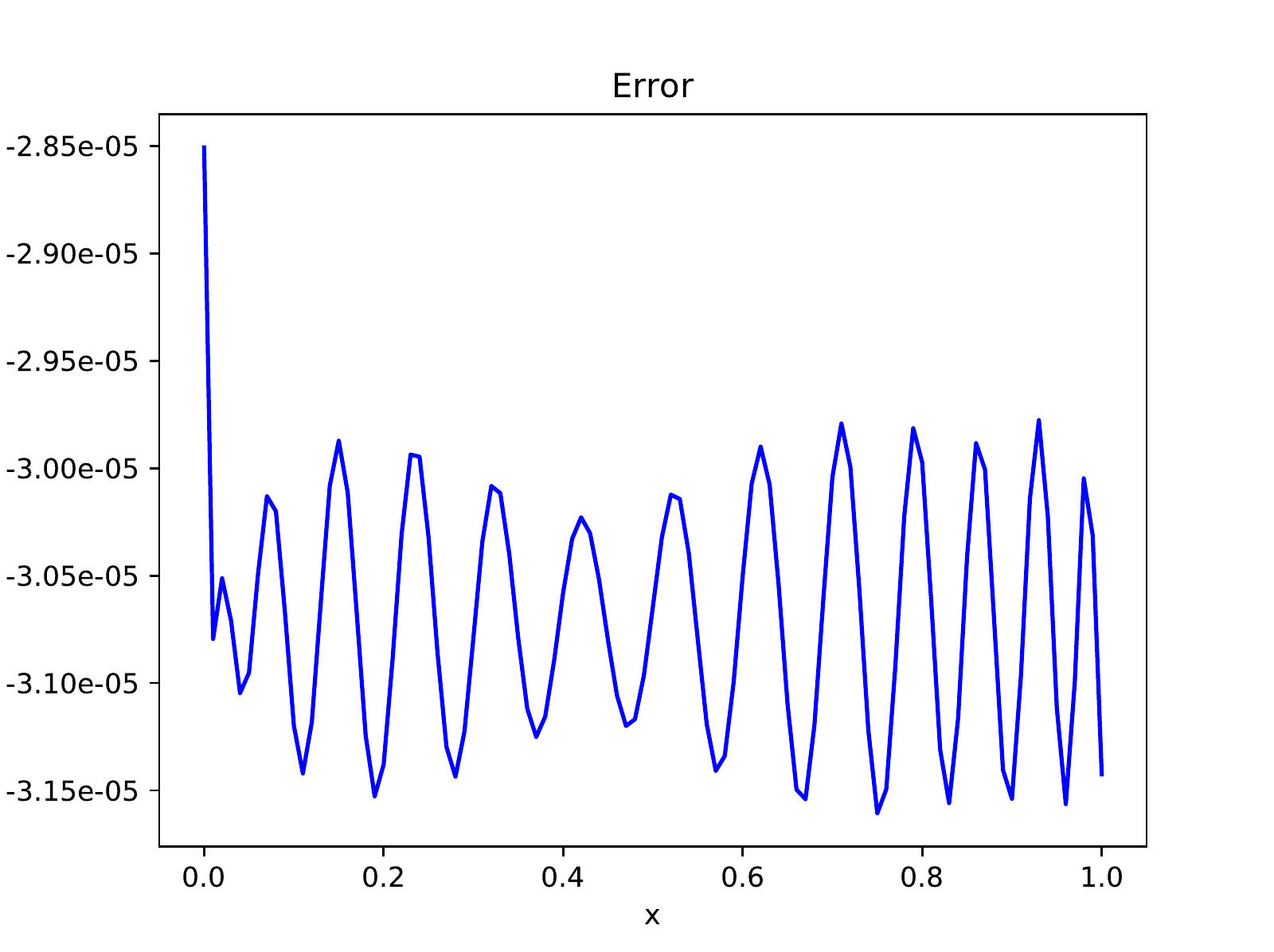}
\caption{The difference between the exact and approximated solution.}
\label{advecerr1d}
\end{subfigure}
\caption{Approximate solution and error. The solution is computed with 100 equidistant collocation points using an ANN with 2 hidden layers with 10 neurons each.}
\label{advecfig1d}
\end{figure}

The stationary, scalar, linear advection equation in 1D is given by
\begin{equation}
\begin{aligned}
Lu &= \frac{du}{dx} = f, && 0 < x \leq 1, \\
u(0) &= g_0.
\end{aligned}
\label{advecex1d}
\end{equation}
To get an analytic solution we take, for example,
\begin{equation}
u = \sin(2\pi x)\cos(4\pi x) + 1
\end{equation}
and plug it into (\ref{advecex1d}) to compute $f$ and $g_0$. In this simple case the boundary data extension can be taken as the constant $G(x) \equiv u(0) = 1$ and the smoothed distance function can be seen in Figure~\ref{distadvec}. In this simple case we could take the distance function to be the line $D(x) = x$. When $L$ acts on the ansatz $\hat{u}$ we get
\begin{equation}
L\hat{u} = \frac{dG}{dx} + \frac{dD}{dx}y^L_1 + D\frac{\partial y^L_1}{\partial x}.
\label{luhatadvec1d}
\end{equation}
To use a gradient based optimization method we need the gradients of the quadratic residual cost function (\ref{rescost}). They can be computed from (\ref{rescostgrad}) as
\begin{equation}
\begin{aligned}
\frac{\partial C}{\partial w^l_{jk}} &= (L \hat{u} - f) \left(\frac{dD}{dx} \frac{\partial y^L_1}{\partial w^l_{jk}} + D \frac{\partial^2y^l_1}{\partial x \partial w^l_{jk}}\right), \\
\frac{\partial C}{\partial b^l_j} &= (L \hat{u} - f) \left(\frac{dD}{dx} \frac{\partial y^L_1}{\partial b^l_j} + D \frac{\partial^2y^l_1}{\partial x \partial b^l_j}\right).
\end{aligned}
\label{rescostadvec1d}
\end{equation}
Each of the terms in (\ref{luhatadvec1d}) and (\ref{rescostadvec1d}) can be computed using the algorithms (\ref{feedforward}), (\ref{backprop}), (\ref{feedforward1}), (\ref{backprop0}), or (\ref{backprop1}).  The result is shown in Figure~\ref{advecfig1d}. In this case we used an ANN with 2 hidden layers with 10 neurons each and 100 equidistant collocation points.

\subsection{Linear diffusion in 1D}
\begin{figure}[H]
\centering
\begin{subfigure}[t]{0.49\textwidth}
\includegraphics[width=\textwidth]{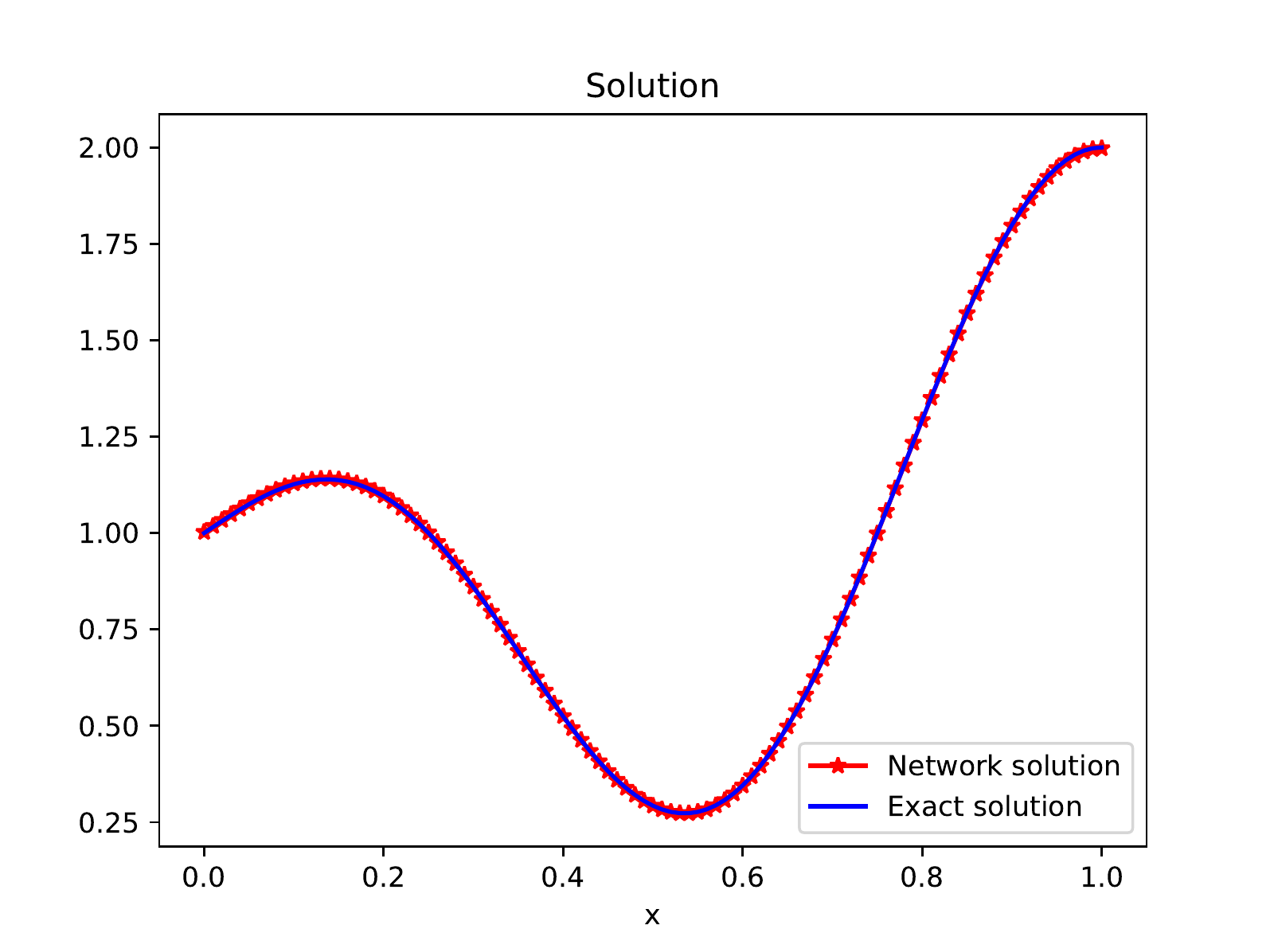}
\caption{Exact and approximate ANN solution to the 1D stationary diffusion equation.}
\label{heatsol1d}
\end{subfigure}
\begin{subfigure}[t]{0.49\textwidth}
\includegraphics[width=\textwidth]{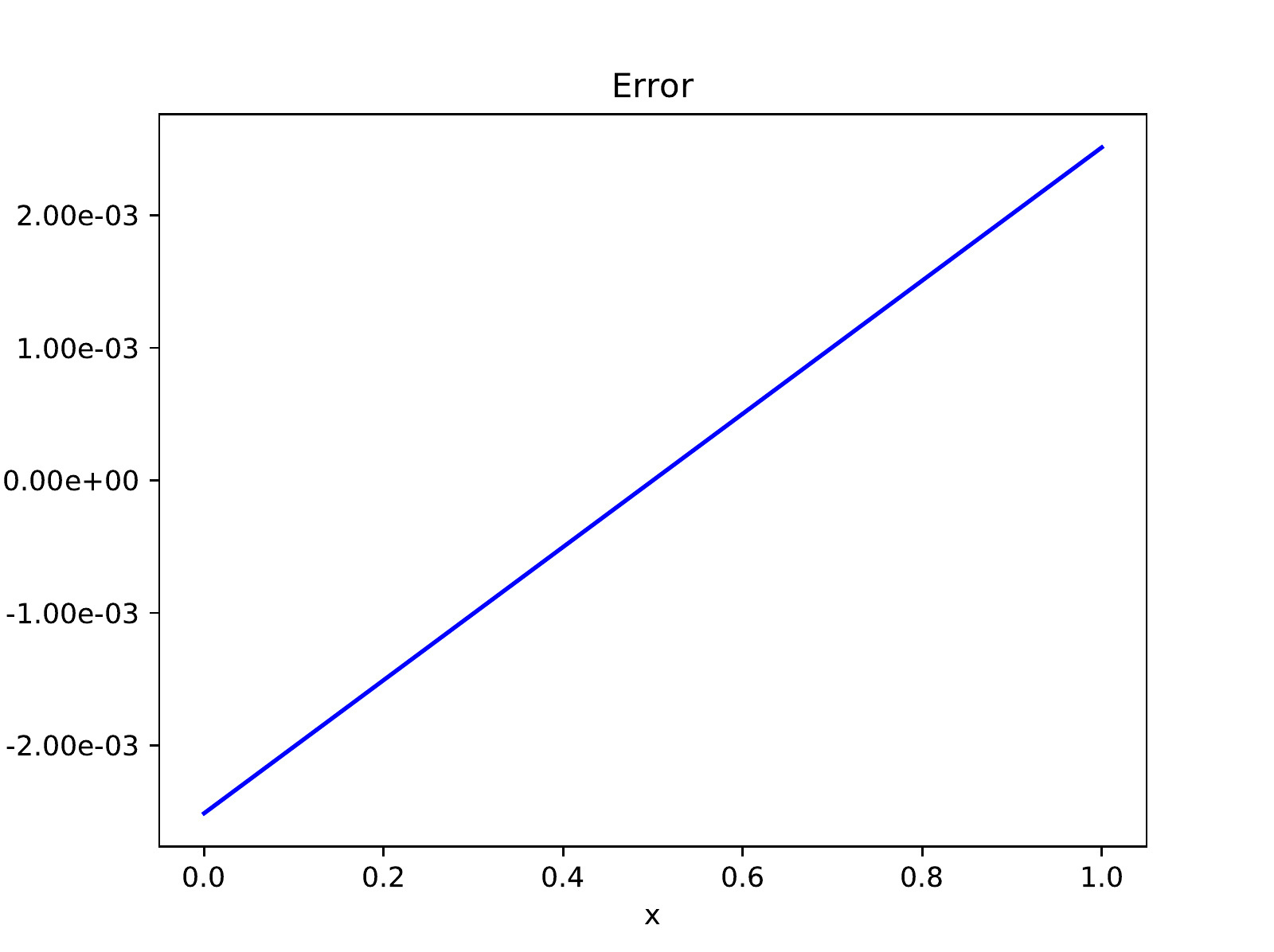}
\caption{The difference between the exact and approximated solution.}
\label{heaterr1d}
\end{subfigure}
\caption{Approximate solution and error. The solution is computed using 100 equidistant collocation points using an ANN with two hidden layers with 10 neurons each.}
\label{heatfig1d}
\end{figure}

The stationary, scalar, linear diffusion equation in 1D is given by
\begin{equation}
\begin{aligned}
Lu &= \frac{d^2u}{dx^2} = f, && 0 < x < 1, \\
u(0) &= g_0, && u(1) = g_1.
\end{aligned}
\end{equation}
To get an analytical solution we let, for example,
\begin{equation}
u = \sin(\frac{\pi x}{2}) \cos(2 \pi x) + 1
\end{equation}
and compute $f$, $g_0$, and $g_1$. In this simple case the boundary data extension can be taken as the straight line $G(x) = x + 1$. The smoothed distance function $D$ is seen in Figure~\ref{distheat}. Here we could instead take the smoothed distance function to be the parabola $D(x) = x(1 - x)$. When $L$ acts on the ansatz we get
\begin{equation}
L\hat{u} = \frac{d^2G}{dx^2} + \frac{d^2D}{dx^2} y^L_1 + 2\frac{dD}{dx} \frac{\partial y^L_1}{\partial x} + D \frac{\partial^2y^L_1}{\partial x^2}
\label{luhatheat1d}
\end{equation}
and to use gradient based optimization we compute (\ref{rescostgrad}) as
\begin{equation}
\begin{aligned}
\frac{\partial C}{\partial w^l_{jk}} &= (L\hat{u} - f) \left( \frac{d^2D}{dx^2} \frac{\partial y^L_1}{\partial w^l_{jk}} + 2\frac{dD}{dx} \frac{\partial^2y^L_1}{\partial x \partial w^l_{jk}} + D \frac{\partial^3y^L_1}{\partial x^2\partial w^l_{jk}} \right), \\
\frac{\partial C}{\partial b^l_j} &= (L\hat{u} - f) \left( \frac{d^2D}{dx^2} \frac{\partial y^L_1}{\partial b^l_j} + 2\frac{dD}{dx} \frac{\partial^2y^L_1}{\partial x \partial b^l_j} + D \frac{\partial^3y^L_1}{\partial x^2\partial b^l_j} \right).
\end{aligned}
\label{rescostheat1d}
\end{equation}
Each of the terms in (\ref{luhatheat1d}) and (\ref{rescostheat1d}) can be computed using the algorithms (\ref{feedforward}), (\ref{feedforward1}), (\ref{backprop0}), (\ref{backprop1}), (\ref{feedforward2}), and (\ref{backprop2}). The solution is shown in Figure~\ref{heatfig1d} using an ANN with two hidden layers with 10 neurons each and 100 equidistant collocation points.

\subsection{A remark on 1D problems}
First and second order problems in 1D have a rather interesting property --- the boundary data can be added in a pure post processing step. A first order problem in 1D requires only one boundary condition and the boundary data extension can be taken as a constant. When the first-order operator acts on the ansatz the boundary data extension vanishes and is not used in any subsequent training. A second order problem in 1D requires boundary data at both boundary points and the boundary data extension can be taken as the line passing through these points. When the second-order operator acts on the ansatz the boundary data extension vanishes and is not used in any subsequent training. For one dimensional problems we hence let the ansatz be given by
\begin{equation}
\tilde{u}(x) = D(x)y^L(x)
\end{equation}
and we can then evaluate the solution for any boundary data using
\begin{equation}
\hat{u}(x) = G(x) + \tilde{u}(x)
\end{equation}
without any re-training. Even 1D equations can be time consuming for complicated problems and large networks and being able to experiment with different boundary data without any re-training is an improvement.

\subsection{Linear advection in 2D}
The stationary, scalar, linear advection equation in 2D is given by
\begin{equation}
\begin{aligned}
Lu  &= a\frac{\partial u}{\partial x} + b\frac{\partial u}{\partial y} = f, && x \in \Omega, \\
u &= g, && x \in \Gamma
\end{aligned}
\label{advec2d}
\end{equation}
where $a$, $b$ are the (constant) advection coefficients. The set $\Gamma \subset \partial \Omega$ is the part of the boundary where boundary conditions should be imposed. Let $n=n(x)$ denote the outer unit normal to $\partial\Omega$ at $x\in \partial\Omega$. Following \cite{energymethod} we have
\begin{equation}
\Gamma=\{x\in \partial\Omega : (a, b) \cdot n(x) <  0\}.
\label{adveccond}
\end{equation}
The condition (\ref{adveccond}), usually called the inflow condition, is easily incorporated into the computation of the distance function. Whenever we generate a boundary point to use in the computation of the distance function we check if (\ref{adveccond}) holds, and if it does not we add the point to the set of collocation points.

\begin{figure}[H]
\centering
\begin{subfigure}[t]{0.49\textwidth}
\centering
\includegraphics[width = \textwidth]{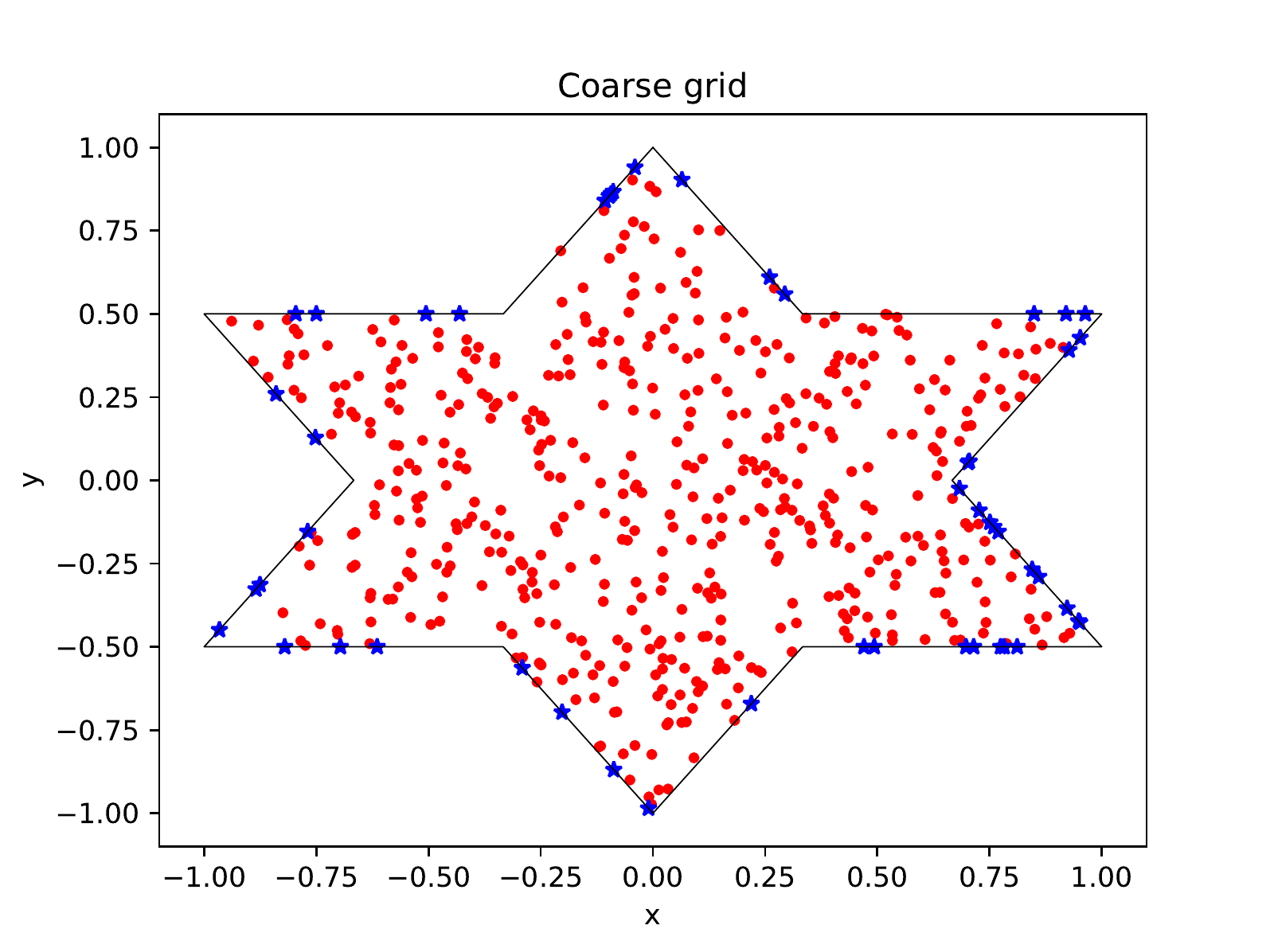}
\caption{$N = 500$ and $M = 50$ for computation of the smoothed distance function.}
\label{stardomaincompute}
\end{subfigure}
\begin{subfigure}[t]{0.49\textwidth}
\centering
\includegraphics[width = \textwidth]{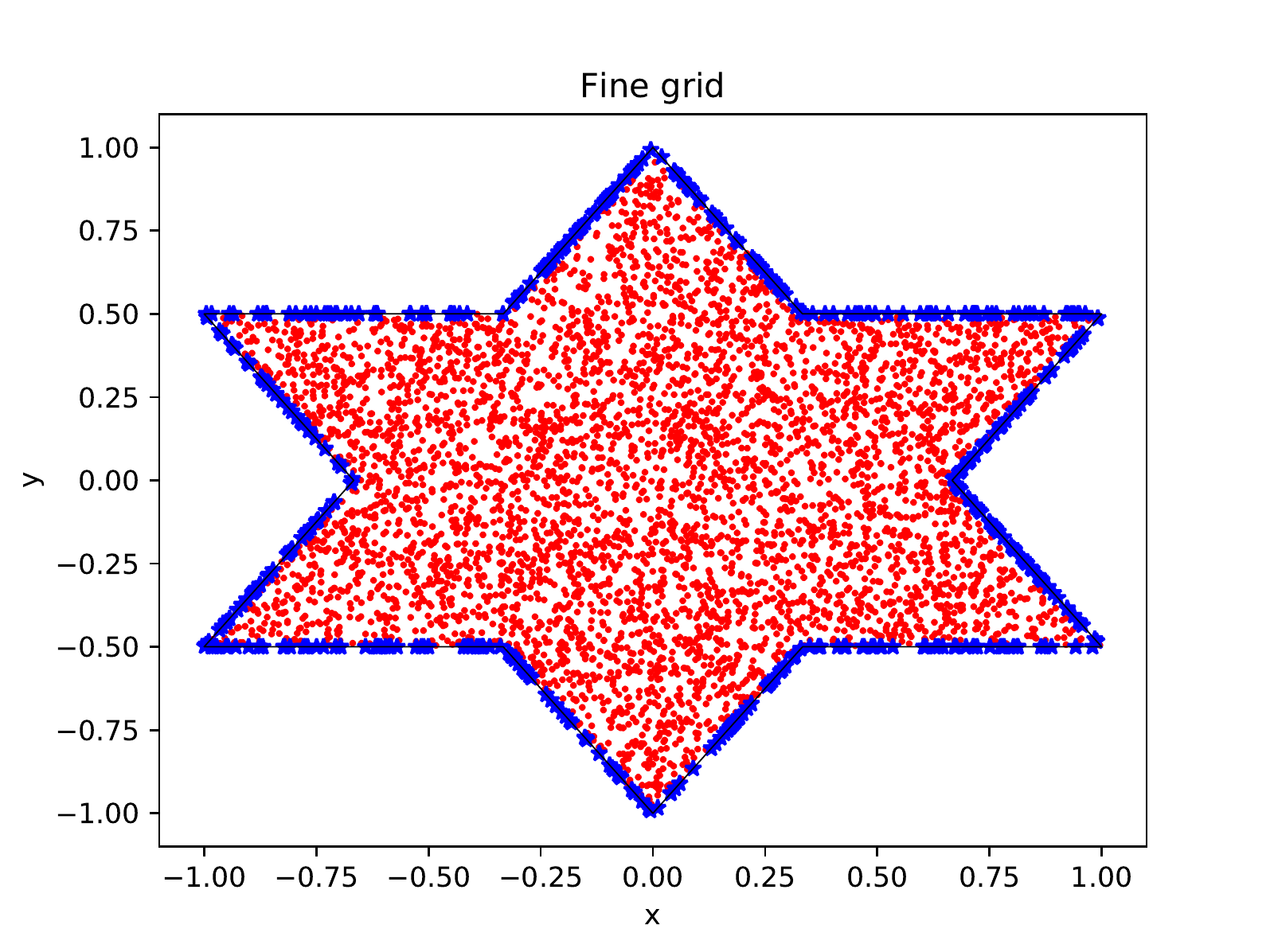}
\caption{$N = 5000$ and $M = 500$ for computation of the boundary data extension and solution.}
\label{stardomaineval}
\end{subfigure}
\caption{Star-shaped domains with uniformly distributed points. The coarse grid is a fixed subset of the fine grid.}
\label{stardomain}
\end{figure}

\begin{figure}[H]
\centering
\begin{subfigure}[t]{0.49\textwidth}
\centering
\includegraphics[width = \textwidth]{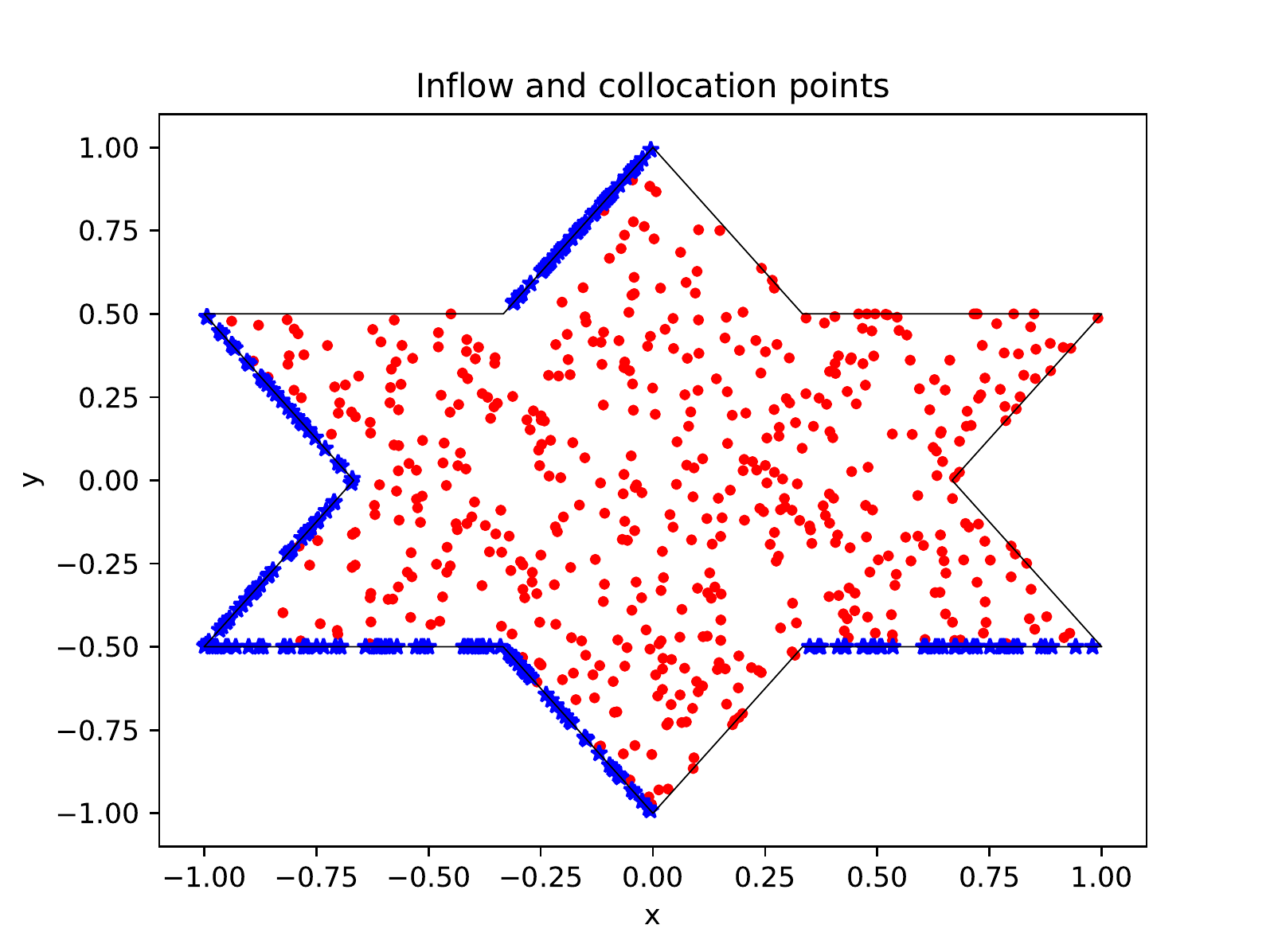}
\caption{Collocations and boundary points satisfying the inflow condition used to compute $d(x)$.}
\label{advecinflow2d}
\end{subfigure}
\begin{subfigure}[t]{0.49\textwidth}
\centering
\includegraphics[width = \textwidth]{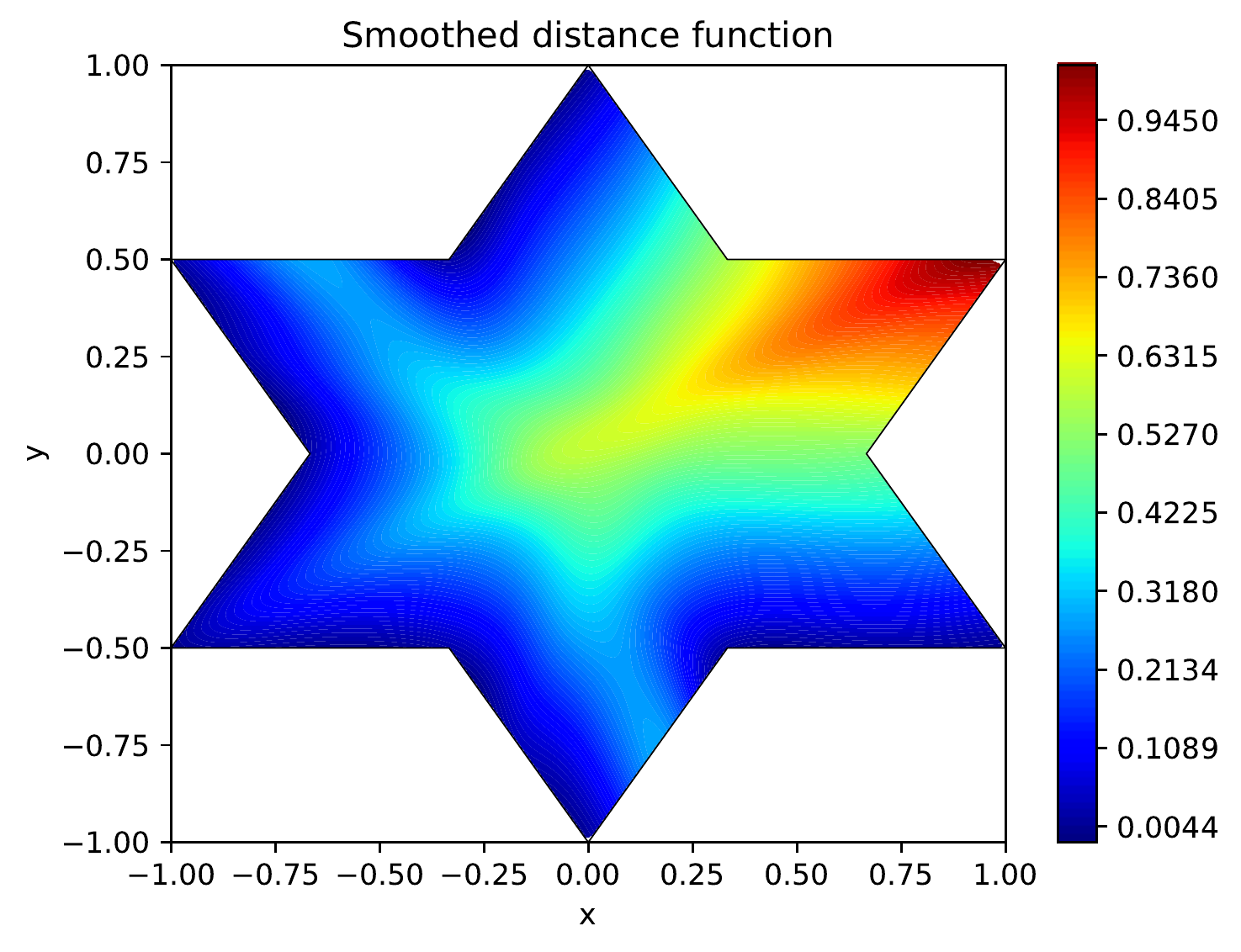}
\caption{Smoothed distance function $D(x)$ computed on the coarse grid using a single hidden layer with 20 neurons.}
\label{advecdistsmooth2d}
\end{subfigure}
\caption{Inflow points and smoothed distance function for the advection equation in 2D with advection coefficients $a = 1$ and $b = 1/2$.}
\label{advecdist2d}
\end{figure}

As an example we take the domain $\Omega$ to be a star shape and generate $N$ uniformly distributed points inside $\Omega$ and $M$ uniformly distributed points on $\partial \Omega$, see Figure~\ref{stardomain}. Once the ANNs have been trained we can evaluate on any number of points.
To compute the distance function we compute (\ref{distfunc}) on the finer grid shown in Figure~\ref{stardomaineval} after traversing the boundary and moving all points that does not satisfy (\ref{adveccond}) to the set of collocation points. To compute the smoothed distance function we train an ANN on the coarser grid shown in Figure~\ref{stardomaincompute} and evaluate it on the finer grid. The results can be seen in Figure~\ref{advecdist2d}.

When the advection operator $L$ acts on the ansatz we get
\begin{equation}
L \hat{u} = a \left( \frac{\partial G}{\partial x} + \frac{\partial D}{\partial x} y^L_1 + D \frac{\partial y^L_1}{\partial x} \right) + b \left(\frac{\partial G}{\partial y} + \frac{\partial D}{\partial y} y^L_1 + D \frac{\partial y^L_1}{\partial y} \right).
\label{luhatadvec2d}
\end{equation}
To compute the solution we need the gradients of the residual cost function given here by
\begin{equation}
\begin{aligned}
\frac{\partial C}{\partial w^l_{ij}} &= (L \hat{u} - f) \left( a \left( \frac{\partial D}{\partial x} \frac{\partial y^L_1}{\partial w^l_{ij}} + D \frac{\partial^2 y^L_1}{\partial x \partial w^l_{ij}} \right) + b \left( \frac{\partial D}{\partial y} \frac{\partial y^L_1}{\partial w^l_{ij}} + D \frac{\partial^2 y^L_1}{\partial y \partial w^l_{ij}} \right) \right), \\
\frac{\partial C}{\partial b^l_j} &= (L \hat{u} - f) \left( a \left( \frac{\partial D}{\partial x} \frac{\partial y^L_1}{\partial b^l_j} + D \frac{\partial^2 y^L_1}{\partial x \partial b^l_j} \right) + b \left( \frac{\partial D}{\partial y} \frac{\partial y^L_1}{\partial b^l_j} + D \frac{\partial^2 y^L_1}{\partial y \partial b^l_j} \right) \right).
\end{aligned}
\label{rescostadvec2d}
\end{equation}
Each of the terms in (\ref{luhatadvec2d}) and (\ref{rescostadvec2d}) can be computed using the algorithms (\ref{feedforward}), (\ref{backprop}), (\ref{feedforward1}), (\ref{backprop0}), or (\ref{backprop1}) as before. Note that the full gradient vectors can be computed simultaneously and there is no need to do a separate feedforward and backpropagation pass for each component of the gradient.

\begin{figure}[H]
\centering
\begin{subfigure}[t]{0.49\textwidth}
\centering
\includegraphics[width = \textwidth]{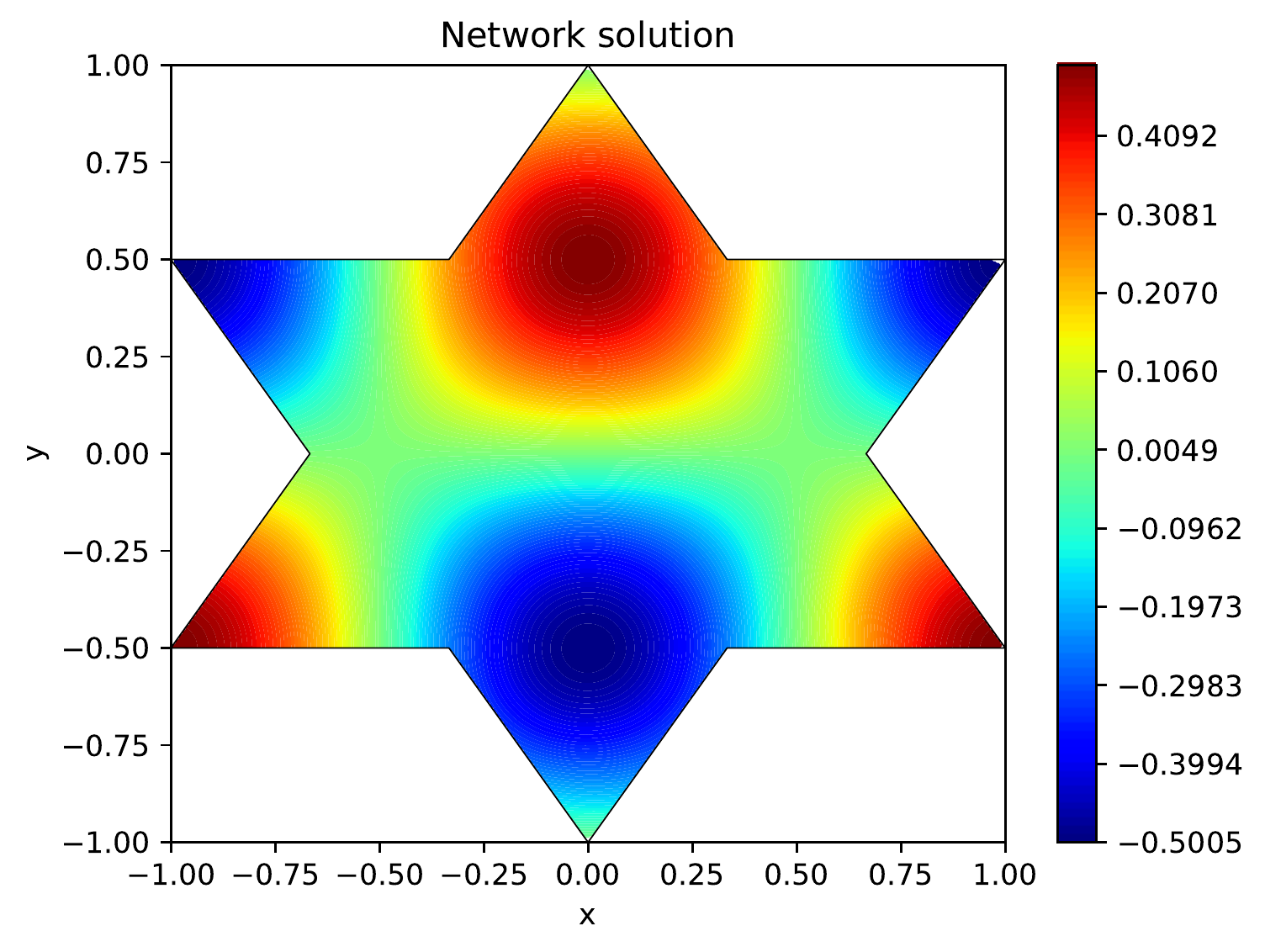}
\caption{ANN solution to the 2D stationary advection equation.}
\label{advecsol2d}
\end{subfigure}
\begin{subfigure}[t]{0.49\textwidth}
\centering
\includegraphics[width = \textwidth]{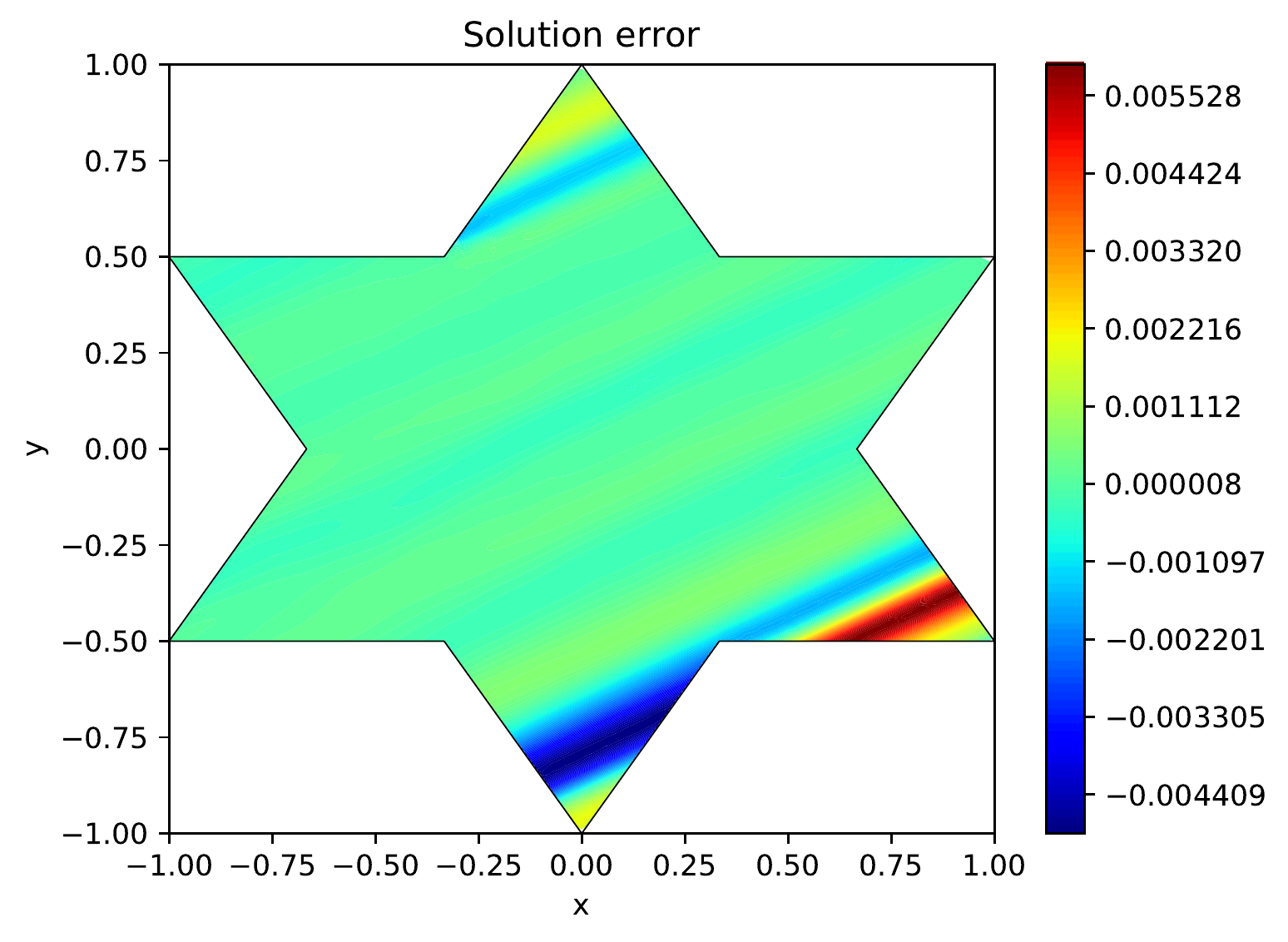}
\caption{Difference between the exact and computed solution.}
\label{advecerr2d}
\end{subfigure}
\caption{Solution and error for the advection equation in 2D with advection coefficients $a = 1$ and $b = 1/2$ using five hidden layers with 10 neurons each.}
\label{advecfig2d}
\end{figure}

To get an analytic solution we let, for example,
\begin{equation}
u = \frac{1}{2} \cos(\pi x) \sin(\pi y)
\end{equation}
and plug it into (\ref{advec2d}) to compute $f$ and $g$. An ANN with a single hidden layer with 20 neurons is used to compute the extension of the boundary data $G$. To compute the solution we use an ANN with five hidden layers with 10 neurons each. The result can be seen in Figure~\ref{advecfig2d}. Note that the errors follow the so-called streamlines. This is a well-known problem in FEM where streamline artificial diffusion is added to reduce the streamline errors \cite{streamline}.

\subsection{Linear diffusion in 2D}
\begin{figure}[H]
\centering
\begin{subfigure}[t]{0.49\textwidth}
\centering
\includegraphics[width = \textwidth]{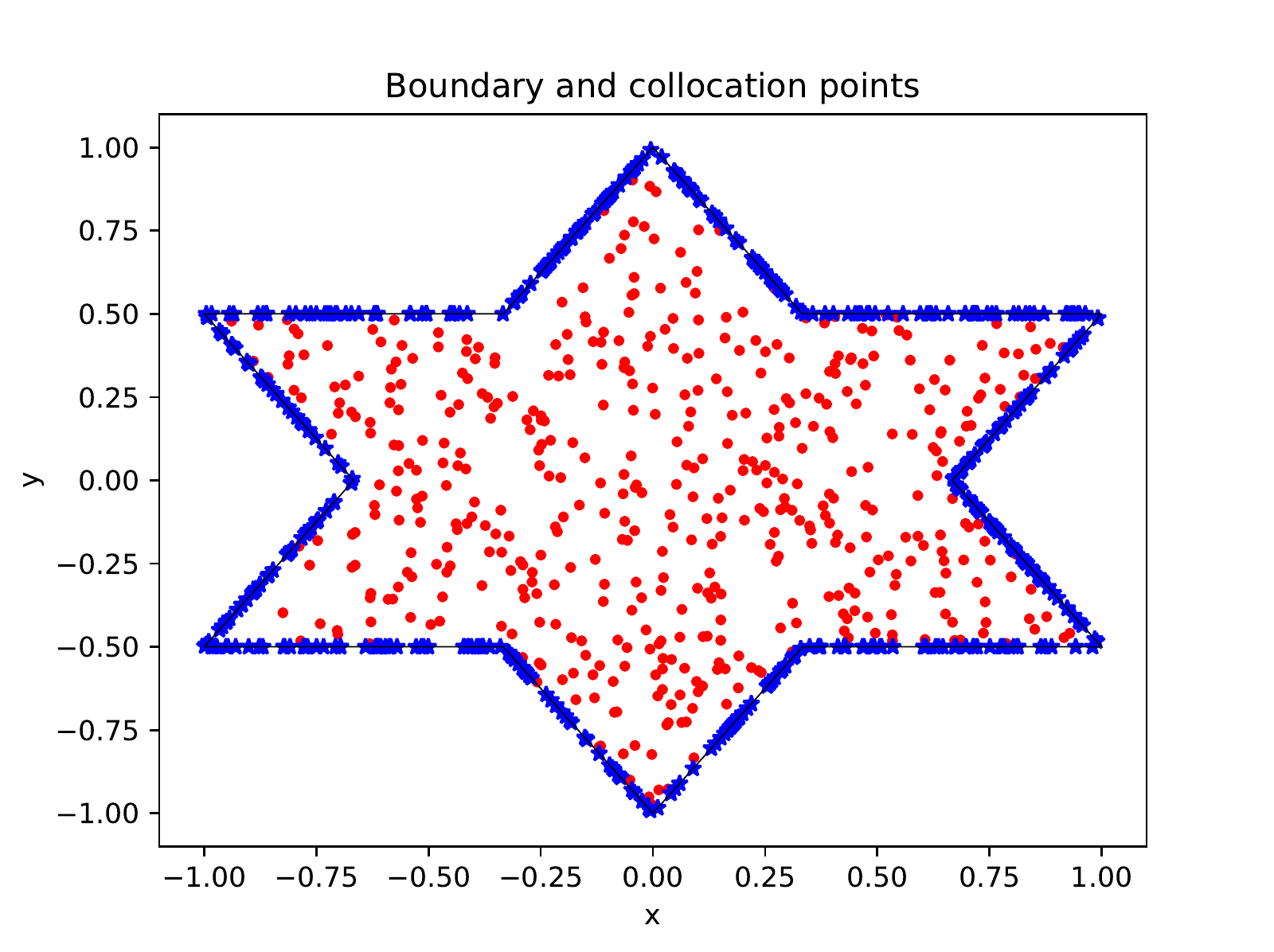}
\caption{Collocation and boundary points used to compute $d(x)$.}
\label{heatinflow2d}
\end{subfigure}
\begin{subfigure}[t]{0.49\textwidth}
\centering
\includegraphics[width = \textwidth]{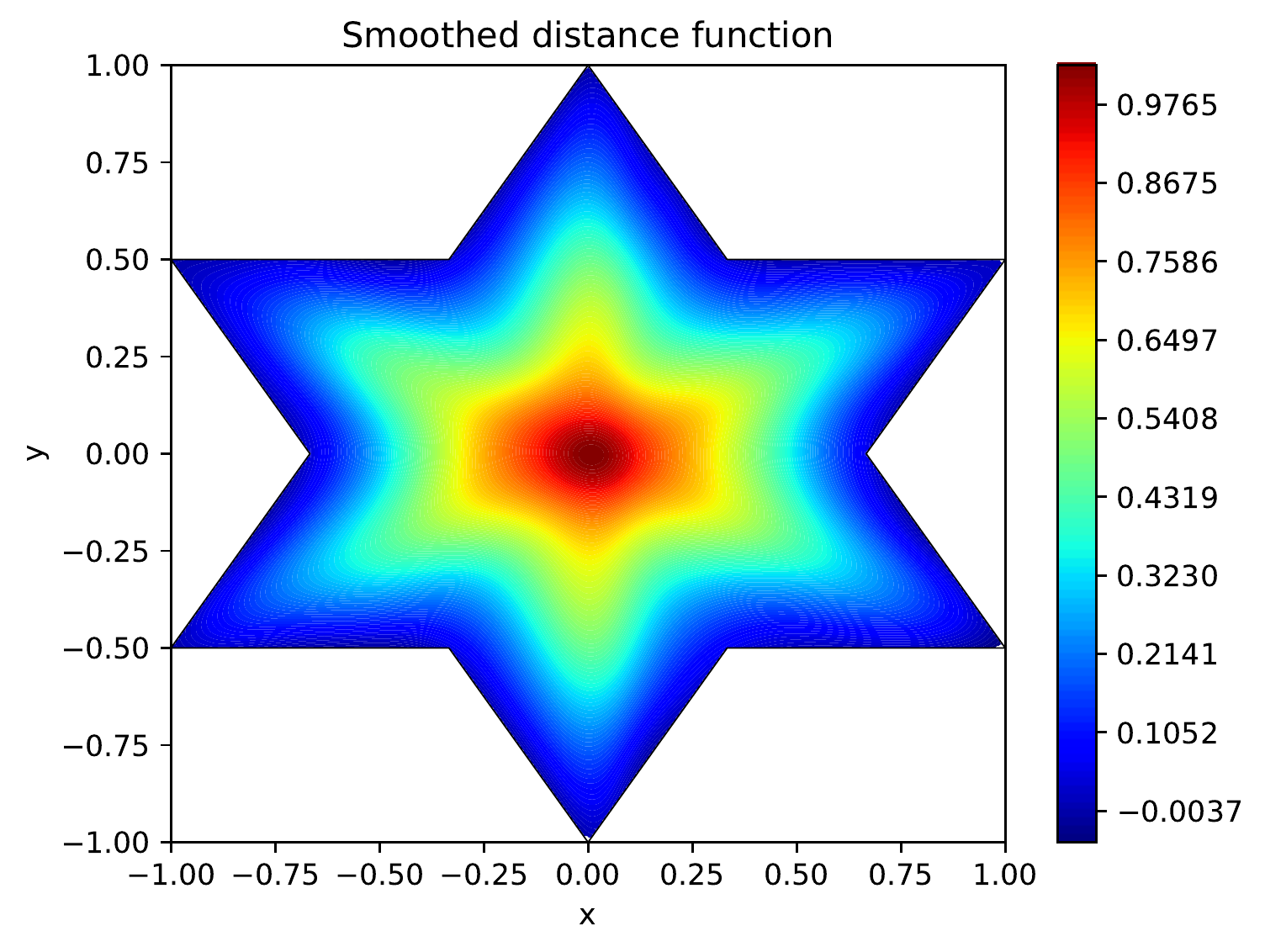}
\caption{Smoothed distance function $D(x)$ computed on the coarse grid using a single hidden layer with 20 neurons.}
\label{heatdistsmooth2d}
\end{subfigure}
\caption{Boundary points and the smoothed distance function for the 2D diffusion equation.}
\label{heatdist2d}
\end{figure}

\begin{figure}[H]
\centering
\begin{subfigure}[t]{0.49\textwidth}
\centering
\includegraphics[width = \textwidth]{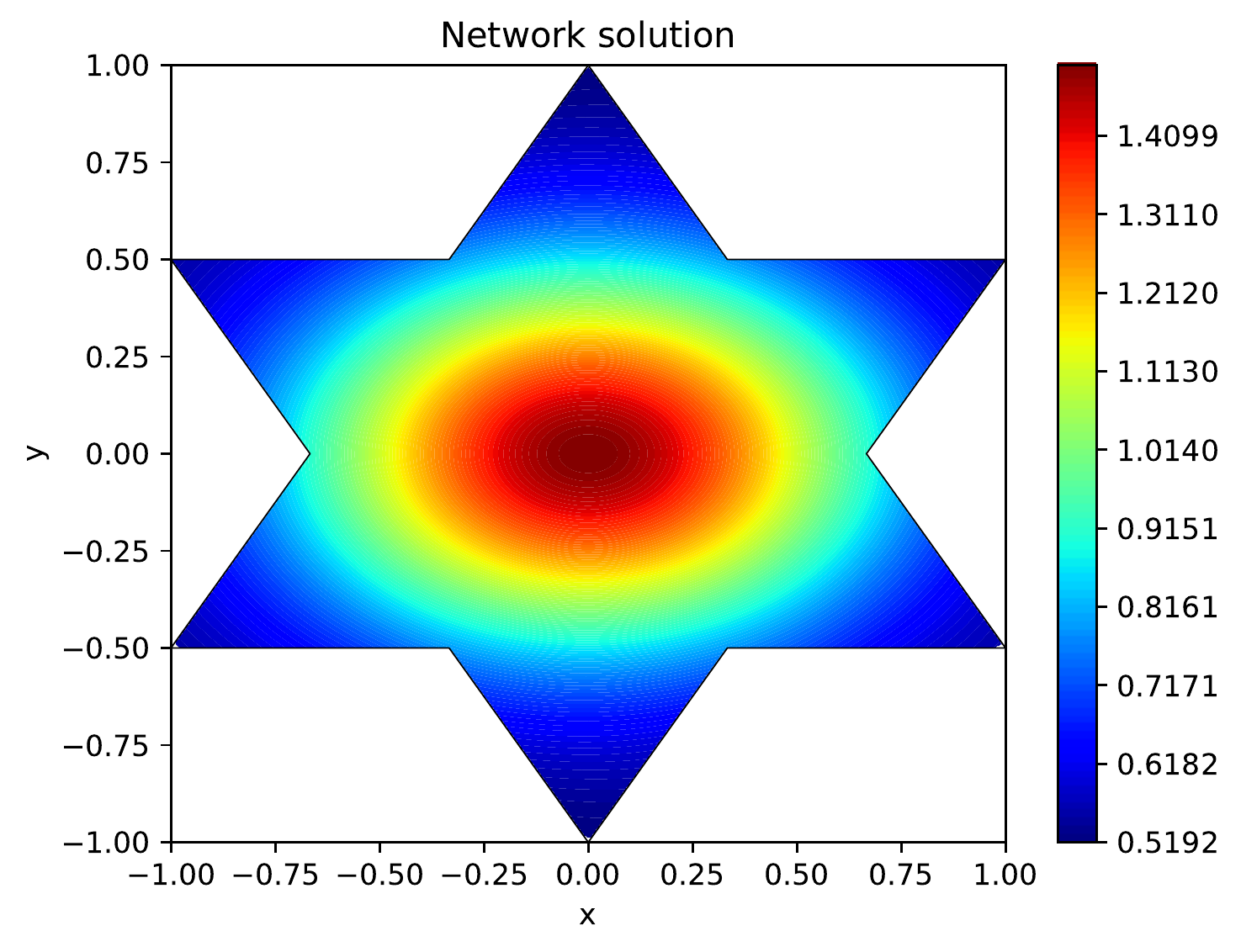}
\caption{ANN solution to the 2D stationary diffusion equation.}
\label{heatsol2d}
\end{subfigure}
\begin{subfigure}[t]{0.49\textwidth}
\centering
\includegraphics[width = \textwidth]{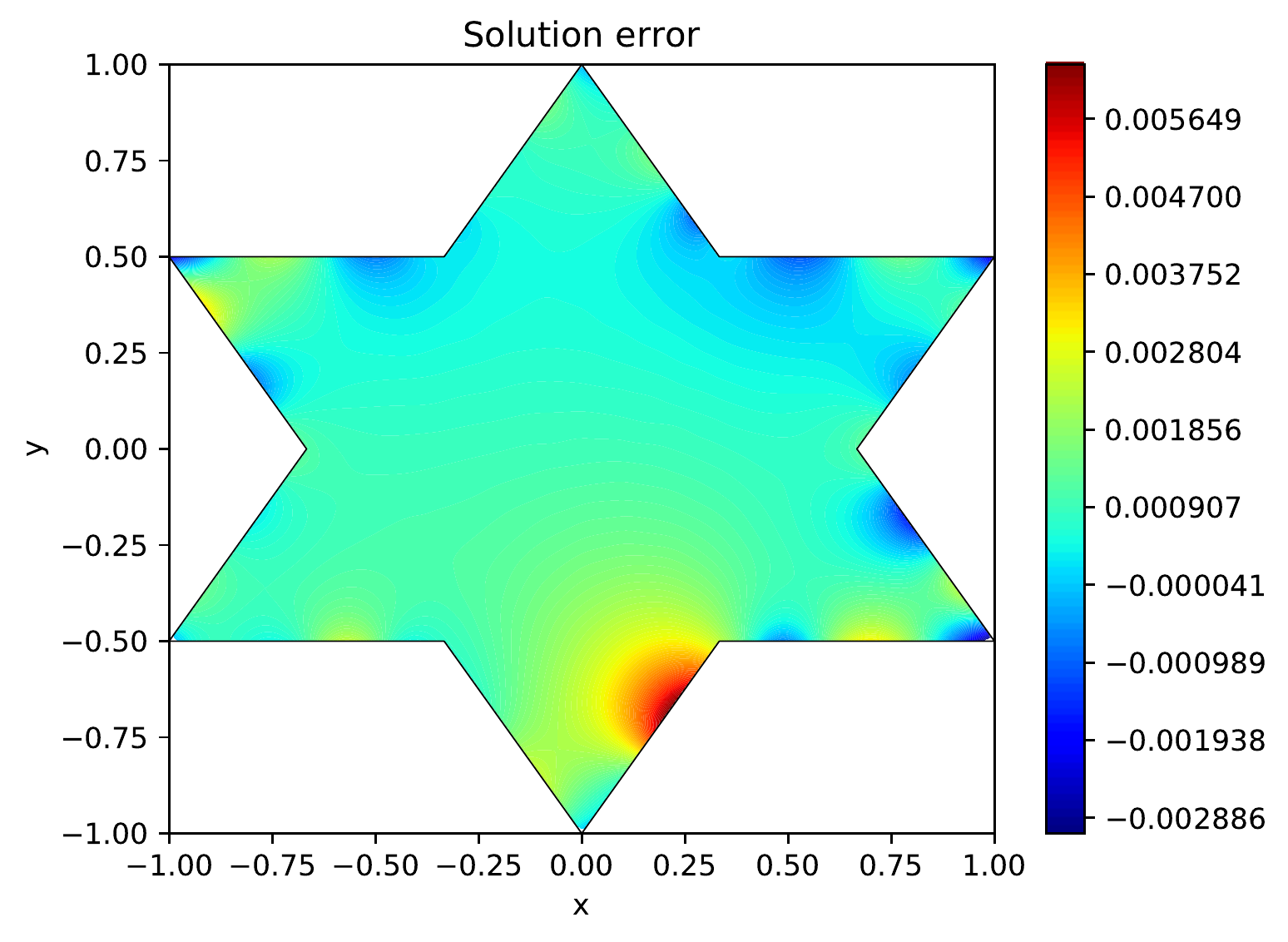}
\caption{Difference between the exact and computed solution.}
\label{heaterr2d}
\end{subfigure}
\caption{Solution and error for the diffusion equation in 2D using five hidden layers with 10 neurons each.}
\label{heatfig2d}
\end{figure}

The linear, scalar diffusion equation in 2D is given by
\begin{equation}
\begin{aligned}
Lu &= \frac{\partial ^2 u}{\partial x^2} + \frac{\partial ^2 u}{\partial y^2} = f, && x \in \Omega, \\
u &= g, && x \in \Gamma.
\end{aligned}
\label{heat2d}
\end{equation}
In this case $\Gamma = \partial \Omega$ and the smoothed distance function is computed using all points along the boundary as seen in Figure~\ref{heatdist2d}.
To get an analytic solution to compare with we let, for example,
\begin{equation}
u = \exp(-(2x^2 + 4y^2)) + \frac{1}{2}
\end{equation}
and use it to compute $f$ and $g$ in (\ref{heat2d}). The extension of  the boundary data is again computed using an ANN with a single hidden layer with 20 neurons and trained on all boundary points of the fine mesh. When the diffusion operator acts on the ansatz we get
\begin{equation}
\begin{aligned}
L \hat{u} &= \frac{\partial^2 G}{\partial x^2} + \frac{\partial^2D}{\partial x^2} y^L_1 + 2\frac{\partial D}{\partial x} \frac{\partial y^L_1}{\partial x} + D \frac{\partial^2 y^L_1}{\partial x^2} \\
&+ \frac{\partial^2 G}{\partial y^2} + \frac{\partial^2D}{\partial y^2} y^L_1 + 2\frac{\partial D}{\partial y} \frac{\partial y^L_1}{\partial y} + D \frac{\partial^2 y^L_1}{\partial y^2},
\end{aligned}
\label{luhatheat2d}
\end{equation}
and the gradients of the residual cost function becomes
\begin{equation}
\begin{aligned}
\frac{\partial C}{\partial w^l_{jk}} &= (L \hat{u} - f) \left( \frac{\partial^2 D}{\partial x^2} \frac{\partial y^L_1}{\partial w^l_{jk}} + 2\frac{\partial D}{\partial x} \frac{\partial^2 y^L_1}{\partial x \partial w^l_{jk}} + D \frac{\partial^3 y^L_1}{\partial x^2 \partial w^l_{jk}} \right) \\
&+ (L \hat{u} - f) \left( \frac{\partial^2 D}{\partial y^2} \frac{\partial y^L_1}{\partial w^l_{jk}} + 2\frac{\partial D}{\partial y} \frac{\partial^2 y^L_1}{\partial y \partial w^l_{jk}} + D \frac{\partial^3 y^L_1}{\partial y^2 \partial w^l_{jk}} \right), \\
\frac{\partial C}{\partial b^l_j} &= (L \hat{u} - f) \left( \frac{\partial^2 D}{\partial x^2} \frac{\partial y^L_1}{\partial b^l_j} + 2\frac{\partial D}{\partial x} \frac{\partial^2 y^L_1}{\partial x \partial b^l_j} + D \frac{\partial^3 y^L_1}{\partial x^2 \partial b^l_j} \right) \\
&+ (L \hat{u} - f) \left( \frac{\partial^2 D}{\partial y^2} \frac{\partial y^L_1}{\partial b^l_j} + 2\frac{\partial D}{\partial y} \frac{\partial^2 y^L_1}{\partial y \partial b^l_j} + D \frac{\partial^3 y^L_1}{\partial y^2 \partial b^l_j} \right).
\end{aligned}
\label{rescostheat2d}
\end{equation}
Each of the terms in (\ref{rescostheat2d}) and (\ref{luhatheat2d}) can be computed using the algorithms (\ref{feedforward}), (\ref{feedforward1}), (\ref{backprop0}), (\ref{backprop1}), (\ref{feedforward2}), and (\ref{backprop2}). The solution using an ANN with five hidden layers with 10 neurons each can be seen in Figure~\ref{heatfig2d}.

\subsection{Linear diffusion in a complex 2D geometry}
\begin{figure}[H]
\centering
\begin{subfigure}[t]{0.402\textwidth}
\centering
\includegraphics[width = \textwidth]{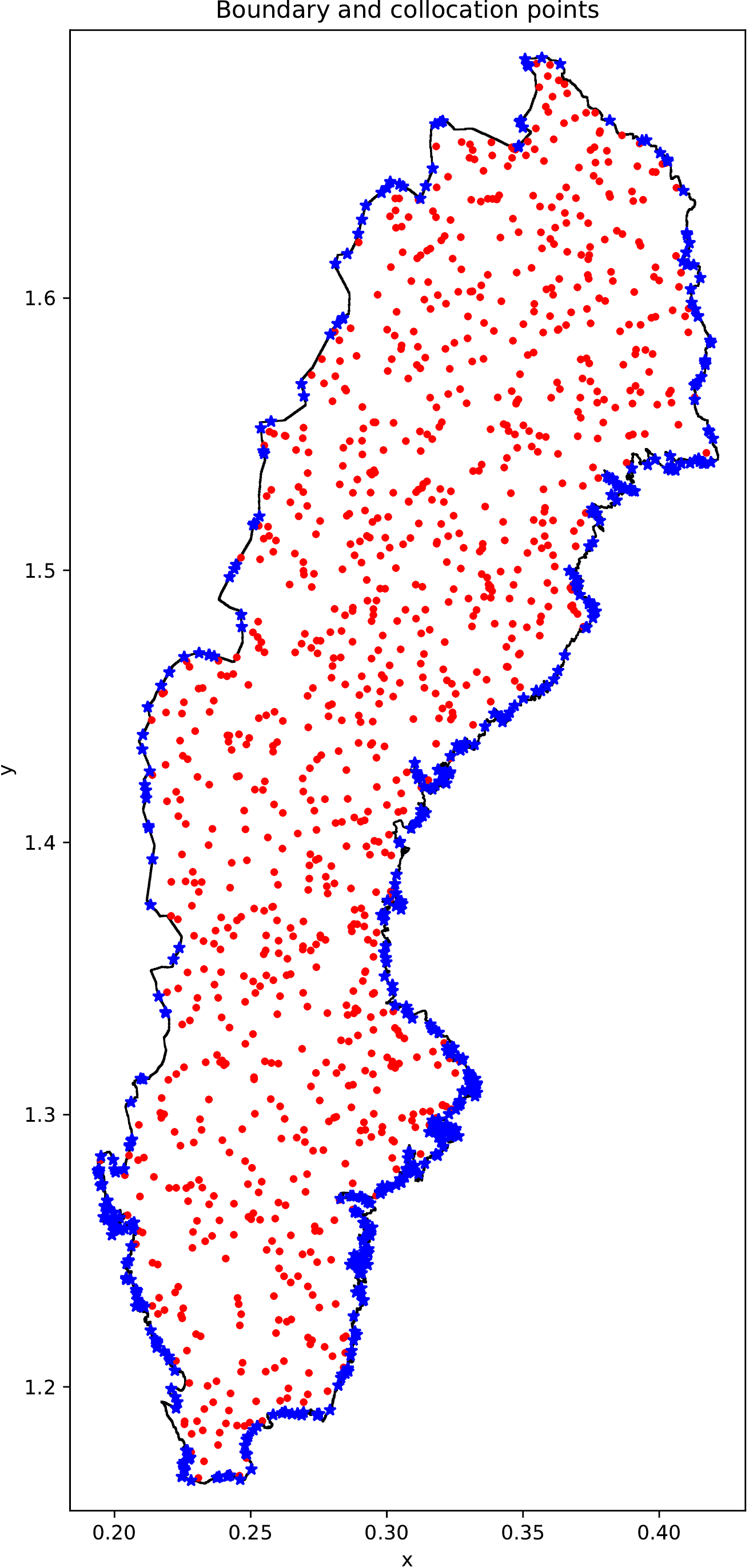}
\caption{Collocation and boundary points used to compute $d(x)$.}
\label{heatinflowsweden}
\end{subfigure}
\begin{subfigure}[t]{0.588\textwidth}
\centering
\includegraphics[width = \textwidth]{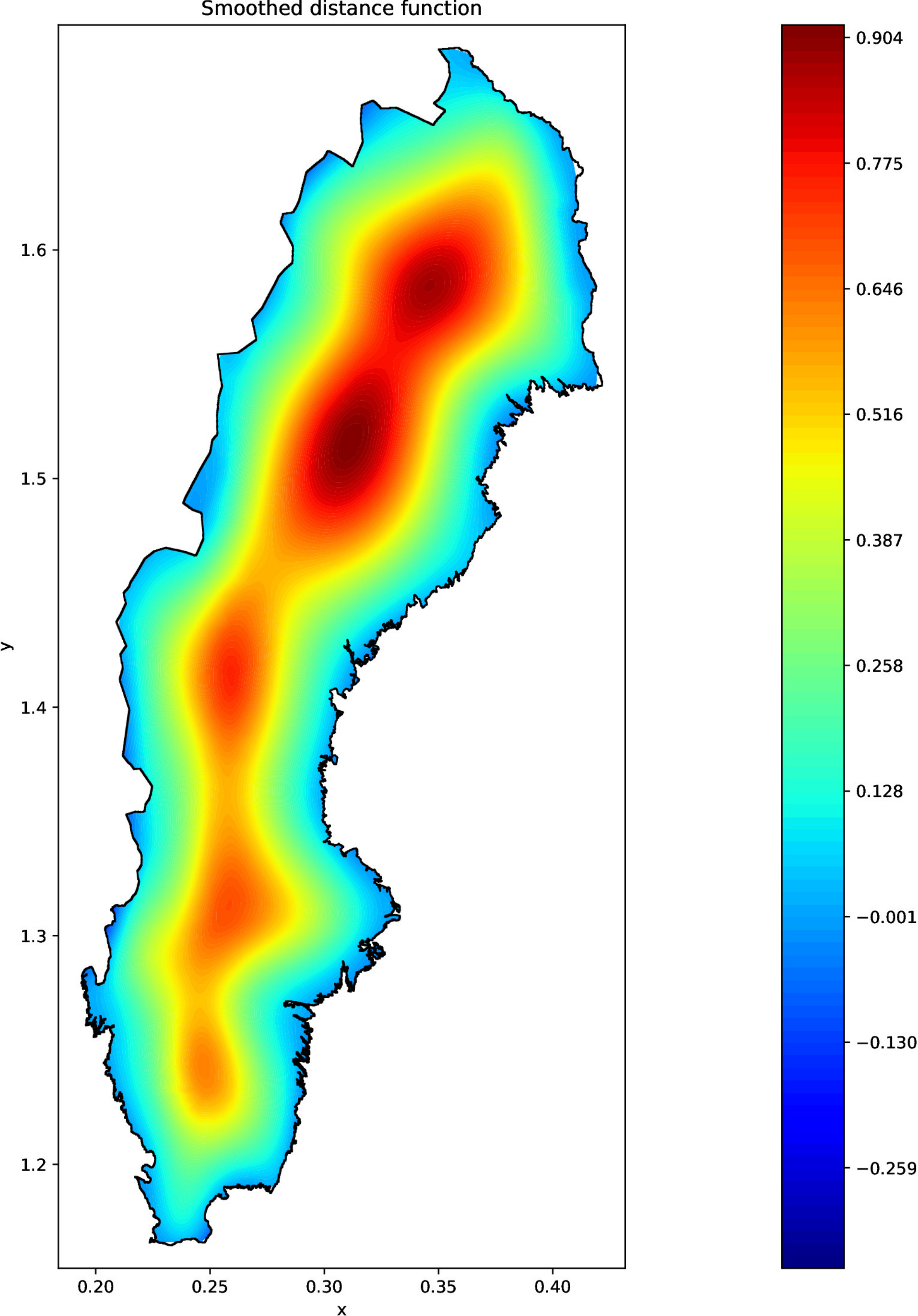}
\caption{Smoothed distance function using a single hidden layer with 20 neurons.}
\label{heatdistsmoothsweden}
\end{subfigure}
\caption{Boundary and collocation points to compute the smoothed distance function and solution.}
\label{heatdistsweden}
\end{figure}

\begin{figure}[H]
\centering
\begin{subfigure}[t]{0.481\textwidth}
\centering
\includegraphics[width = \textwidth]{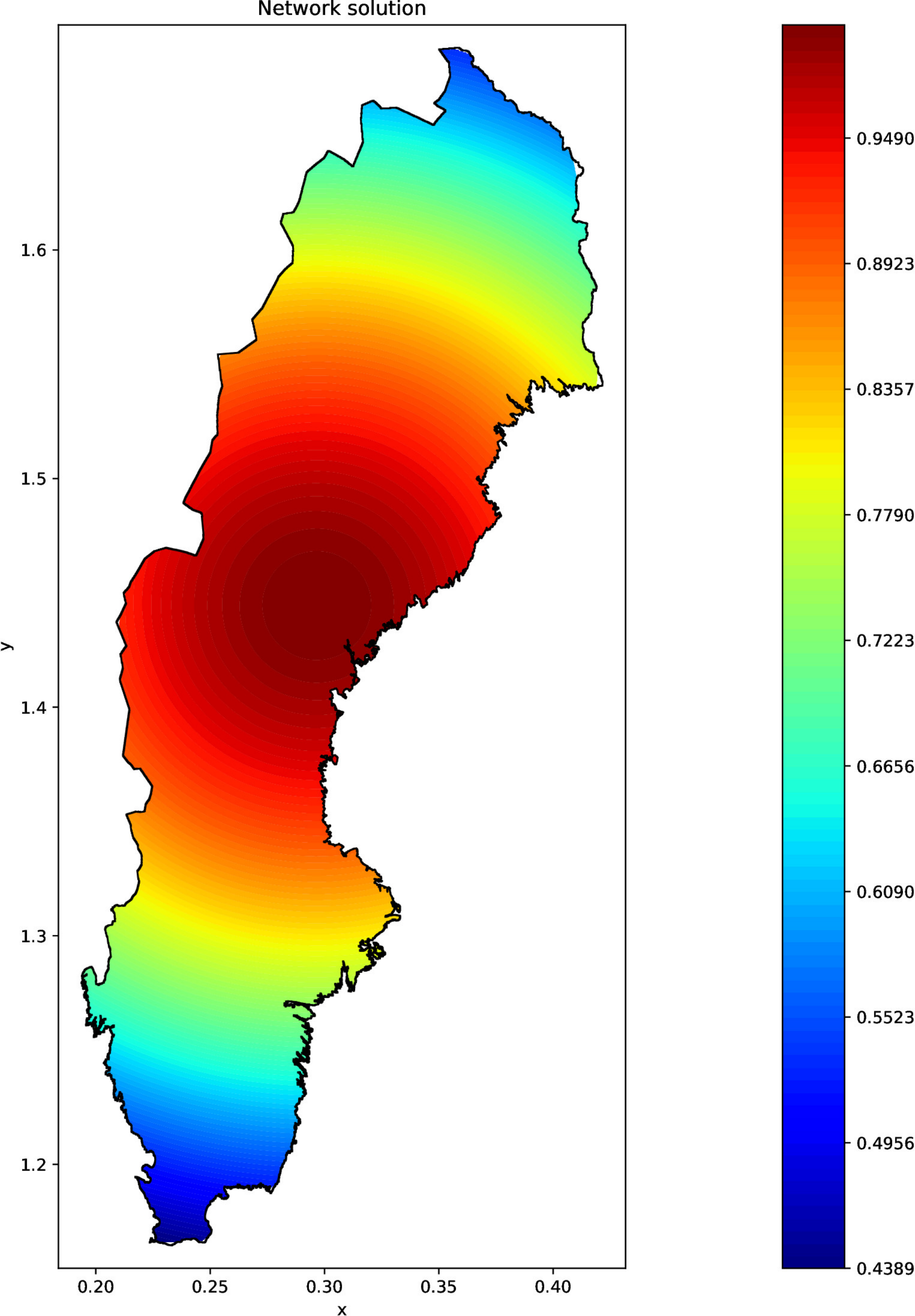}
\caption{ANN solution to the 2D stationary diffusion equation.}
\label{heatsolsweden}
\end{subfigure}
\begin{subfigure}[t]{0.499\textwidth}
\centering
\includegraphics[width = \textwidth]{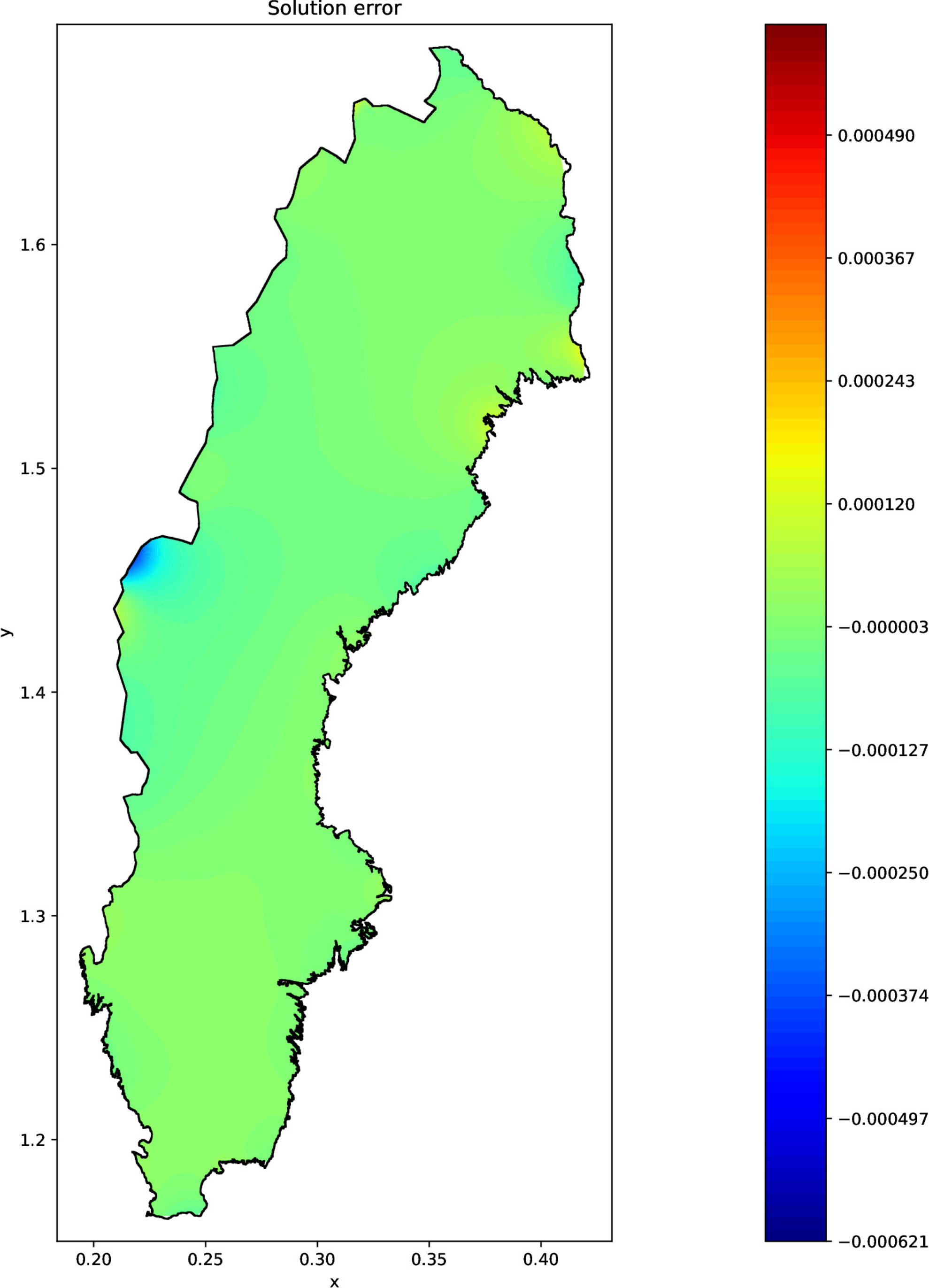}
\caption{Difference between the exact and computed solution.}
\label{heaterrsweden}
\end{subfigure}
\caption{Solution and error for the diffusion equation in a complex 2D geometry using five hidden layers with 10 neurons each.}
\label{heatfigsweden}
\end{figure}

The examples above show how to use the ANN method on non-trivial geometries. To put the method to the test we consider the whole of Sweden as our domain. The domain is represented as a polygon with 160876 vertices and the boundary is highly irregular with very small details. The latitude and longitude coordinates of the verticies have been transformed to $xy$-coordinates by using the standard Mercator projection. As before we generate 500 collocation points uniformly distributed along the boundary and 1000 collocation points inside the domain as shown in Figure~\ref{heatdistsweden} and compute the smoothed distance function using a single hidden layer ANN with 20 neurons. In this case we used all collocation points, rather than a subset as before, to compute the smoothed distance function due to the complexity of the boundary.
We take as an analytic solution the function
\begin{equation}
u = \exp(-10((x - m_x)^2 + (y - m_y)^2))
\end{equation}
where $(m_x, m_y)$ is the center of mass of the domain. The network solution and error can be seen in Figure~\ref{heatfigsweden} where we have as before used a network with 5 hidden layers with 10 neurons each.

\subsubsection{A remark on mesh based methods}
Classical methods for PDEs in complex geometries such as FEM and FVM require that the domain is discretized by a triangulation. Each mesh element along the boundary needs to coincide with a line segment of the polygon describing the boundary. For problems with simple polygonal boundaries for which a high quality mesh can be constructed the ANN method is not competitive with classical methods. Computing gradients and performing gradient based optimization is much more time consuming than solving the linear systems produced by classical numerical methods.

The problem above consists of a complicated polygon with very short line segments and fine grained details which places severe restrictions on the triangulation. In an attempt to compare the ANN method with FEM we used the state of the art FEM software suite \texttt{FEniCS} \cite{LoggMardalEtAl2012a}. \texttt{FEniCS} includes a mesh generator called \texttt{mshr} which is capable of creating high quality meshes on many complex geometries. Unfortunately though, we did not succeed in creating a mesh to be used in a FEM simulation. We experimented with different settings and resolutions but each attempt was aborted after 16 hours when the mesh generation was not completed. The above ANN simulation took about 10 minutes on a high-end laptop from start to finish including computing the smoothed distance function and boundary data extension. With the recent progress in open source software for neural networks by for example TensorFlow \cite{tensorflow} and PyTorch \cite{pytorch}, a reimplementation would significantly speed up the computational time.

\subsection{Higher-dimensional problems}
Higher-dimensional problems follow the same pattern as the 2D case. For example in $N$ dimensions we have the diffusion operator acting on the ansatz as
\begin{equation}
L \hat{u} = \sum_{i = 1}^N \left( \frac{\partial^2 G}{\partial x_i^2} + \frac{\partial^2 D}{\partial x_i^2}y^L_1 + 2 \frac{\partial D}{\partial x_i} \frac{\partial y^L_1}{\partial x_i} + D \frac{\partial y^L_1}{\partial x_i^2} \right),
\end{equation}
and the gradients of the residual cost function becomes
\begin{equation}
\begin{aligned}
\frac{\partial C}{\partial w^l_{jk}} &= (L \hat{u} - f) \sum_{i = 1}^N \left( \frac{\partial^2 D}{\partial x_i^2} \frac{\partial y^L_1}{\partial w^l_{jk}} + 2\frac{\partial D}{\partial x_i} \frac{\partial^2 y^L_1}{\partial x_i \partial w^l_{jk}} + D \frac{\partial^3 y^L_1}{\partial x_i^2 \partial w^l_{jk}} \right), \\
\frac{\partial C}{\partial b^l_j} &= (L \hat{u} - f) \sum_{i = 1}^N \left( \frac{\partial^2 D}{\partial x_i^2} \frac{\partial y^L_1}{\partial b^l_j} + 2\frac{\partial D}{\partial x_i} \frac{\partial^2 y^L_1}{\partial x_i \partial b^l_j} + D \frac{\partial^3 y^L_1}{\partial x_i^2 \partial b^l_j} \right).
\end{aligned}
\end{equation}
As we can see the computational complexity increases linearly with the number of space dimensions as there is one additional term added for each dimension. The number of parameters in the deep ANN increases with the number of neurons in the first hidden layer for each dimension. For a sufficiently deep ANN this increase is negligible.

The main cause for increased computational time is the number of collocation points. Regular grids grows exponentially with the number of dimensions and quickly becomes infeasible. Uniformly random numbers in high-dimensions are in fact not uniformly distributed across the whole space but tends to cluster on hyperplanes, thus providing a poor coverage of the whole space. This is a well-known fact for Monto Carlo methods and has led to the development of sequences of quasi-random numbers called low discrepancy sequences. See for example ch. 2 in \cite{seydel2004tools}, or \cite{2013arXiv1301.3841C, lowdiscrepdist} and the references therein.

There are plenty of different low discrepancy sequences and an evaluation of their efficiency in the context of deep ANNs is beyond the scope of this paper. A common choice is the Sobol sequence \cite{SOBOL196786} which is described in detail in \cite{sobol40} for $N \leq 40$ and later extended in \cite{sobol1111} and \cite{sobol21201} for $N \leq 1111$ and $N \leq 21201$, respectively. To compute a high-dimensional problem we thus generate $N_d$ points from a Sobol sequence in $N$ dimensions to use as collocation points and $N_b$ points in $N - 1$ dimensions to use as boundary points. The rest follows the same procedure as previously described.

A few more examples of work on high-dimensional problems and deep neural networks for PDEs can be found in \cite{dgm, deepcurse, deephigh}.

\section{Convergence considerations}\label{sec5}
Training neural networks consumes a lot of time and computational resources. It is therefore desirable to reduce the number of required iterations until a certain accuracy has been reached. In the following we will discuss two factors which strongly influence the number of required iterations. The first is pre-training the network using the available boundary data and the second is to increase the number of hidden layers.

\subsection{Boundary data pre-training}
\begin{figure}[H]
\centering
\begin{subfigure}[t]{0.49\textwidth}
\centering
\includegraphics[width = \textwidth]{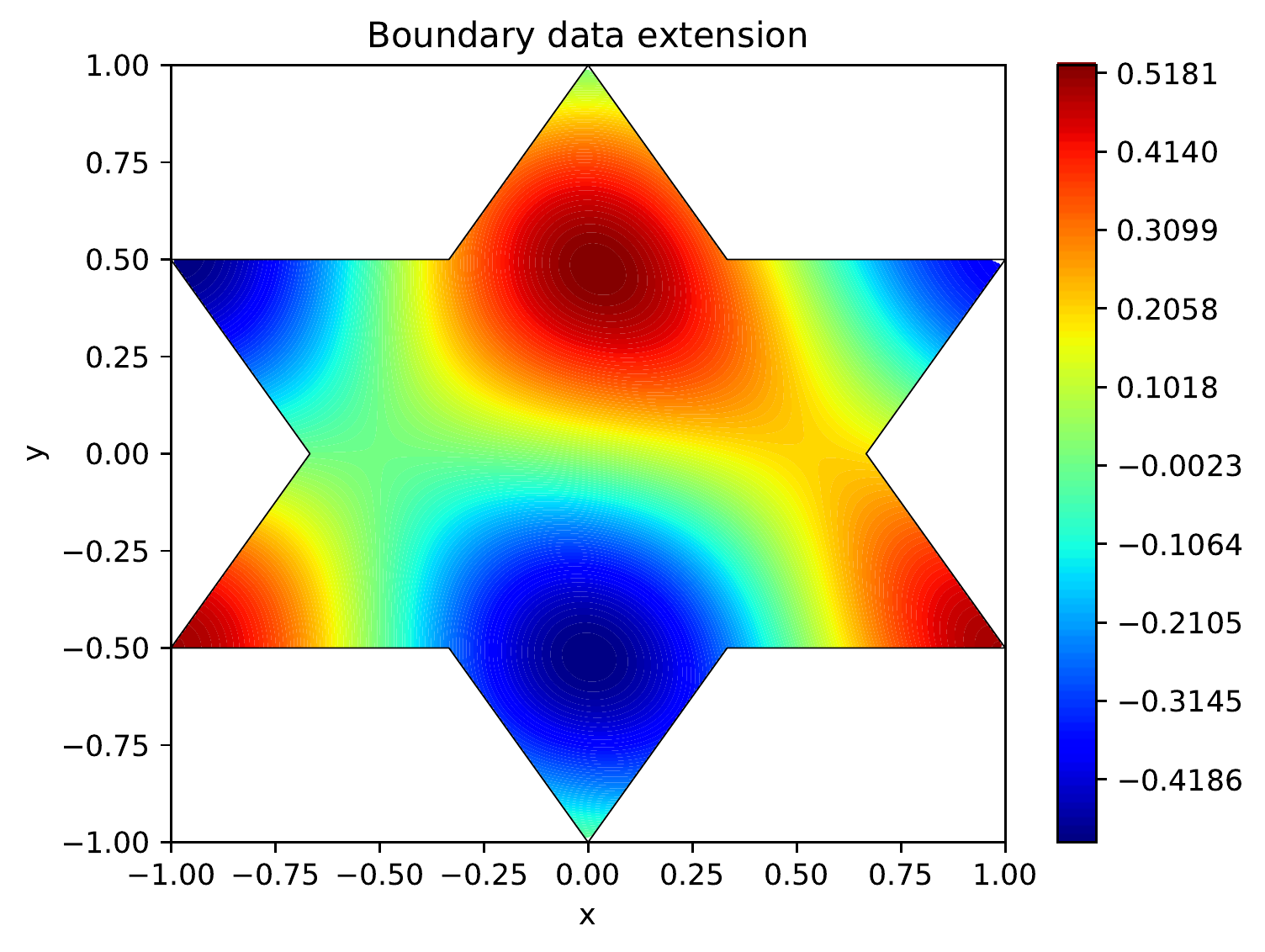}
\caption{Boundary data extension for the advection equation.}
\end{subfigure}
\begin{subfigure}[t]{0.49\textwidth}
\centering
\includegraphics[width = \textwidth]{advecsolnet_2d.pdf}
\caption{Network solution of the advection equation.}
\end{subfigure}
\begin{subfigure}[t]{0.49\textwidth}
\centering
\includegraphics[width = \textwidth]{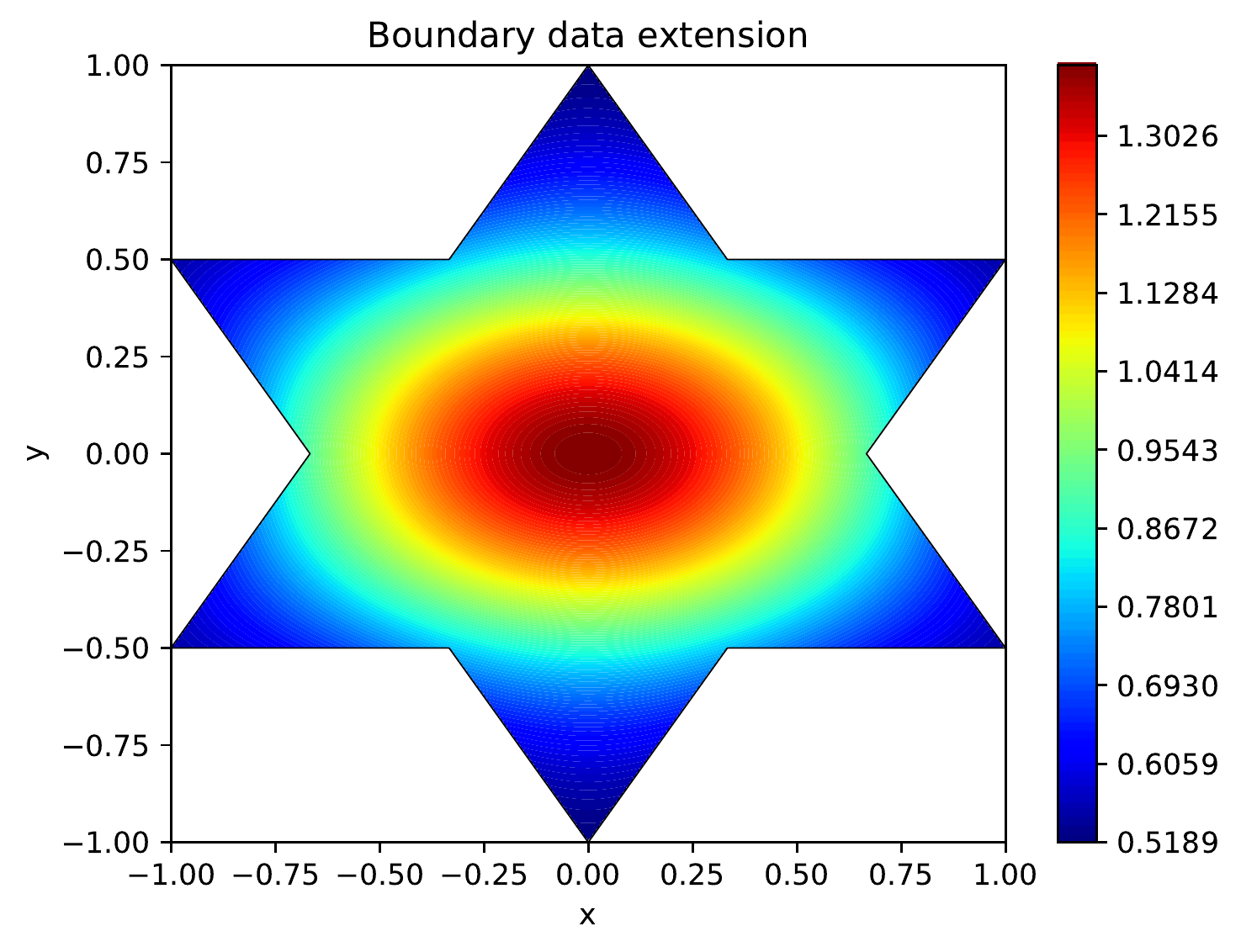}
\caption{Boundary data extension for the diffusion equation.}
\end{subfigure}
\begin{subfigure}[t]{0.49\textwidth}
\centering
\includegraphics[width = \textwidth]{heatsolnet_2d.pdf}
\caption{Network solution of the diffusion equation.}
\end{subfigure}
\caption{The boundary ANNs use a single hidden layer with 20 neurons trained only on the boundary points. The solution ANNs use five hidden layers with 10 neurons each trained on both the boundary and collocation points.}
\label{bndsollowcap}
\end{figure}

As previously described, we used a low-capacity ANN to compute the global extension of the boundary data to save some computational time. However, the low-capacity ANN that extends the boundary data is already an approximate solution to the PDE. Figure~\ref{bndsollowcap} shows the boundary and solution ANNs evaluated on all collocation and boundary points. Note that the boundary ANN has not been trained on a single collocation point inside the domain.

As fitting an ANN to the boundary data is an order of magnitude faster than solving the PDE this suggests that we can pre-train the solution ANN by fitting it to the boundary data. It is still a good idea to keep the boundary ANN as a  separate low-capacity ANN as it will speed up the training of the solution ANN due to the many evaluations that is required during training.

\begin{figure}[H]
\centering
\begin{subfigure}[t]{0.49\textwidth}
\centering
\includegraphics[width = \textwidth]{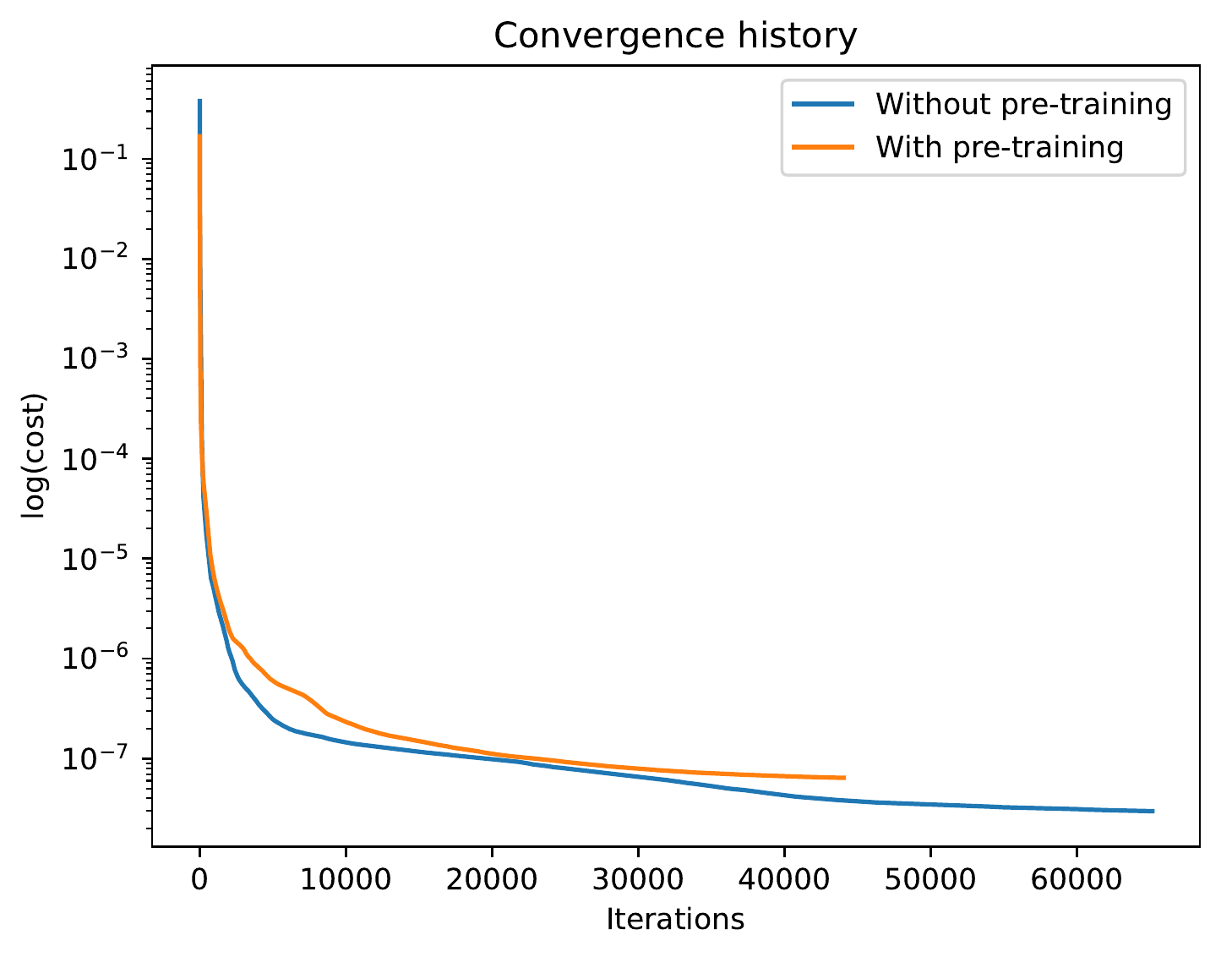}
\caption{Convergence history for the advection equation.}
\end{subfigure}
\begin{subfigure}[t]{0.49\textwidth}
\centering
\includegraphics[width = \textwidth]{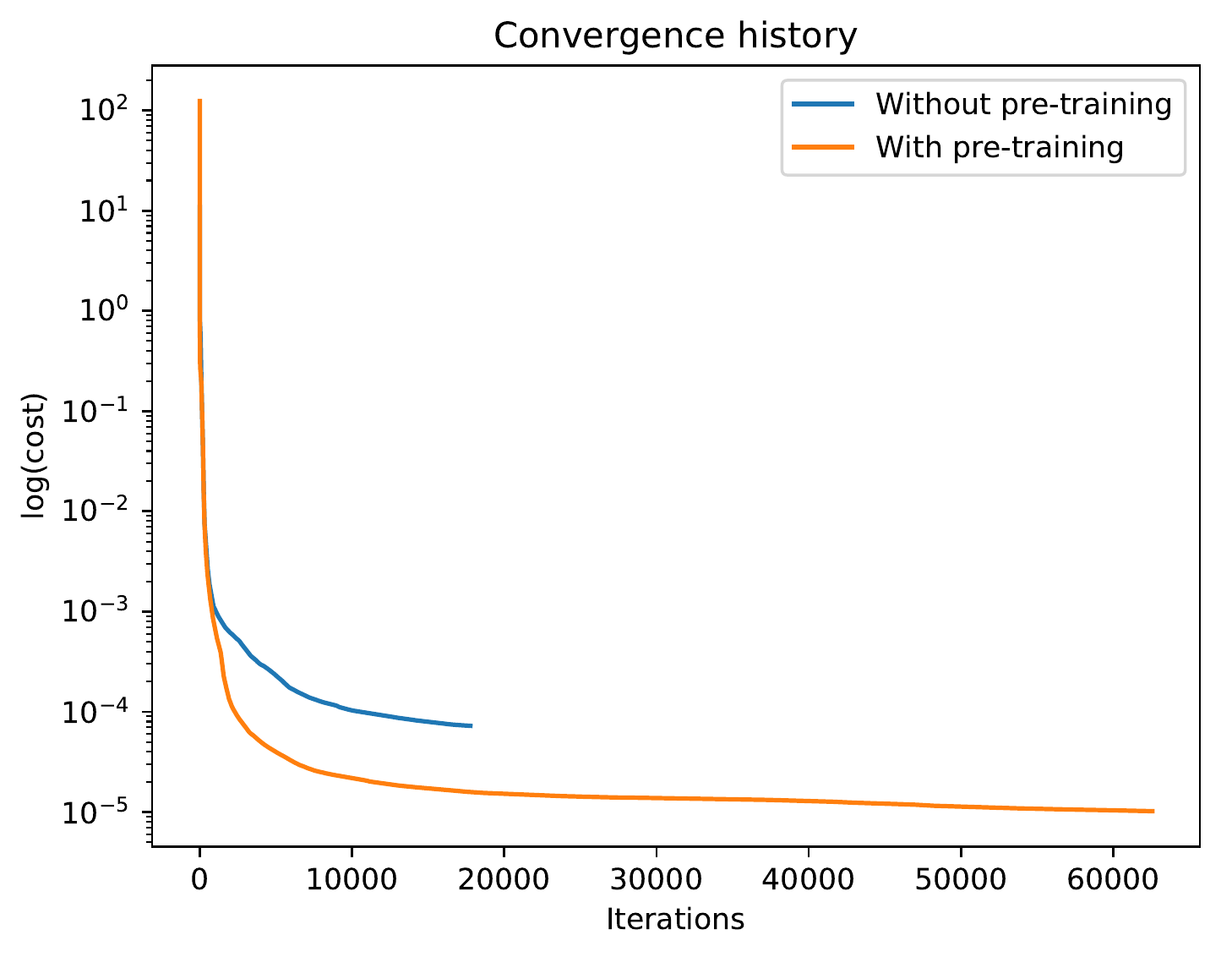}
\caption{Convergence history for the diffusion equation.}
\end{subfigure}
\caption{Convergence history of BFGS until the first line search fail with and without boundary data pre-training.}
\label{convall}
\end{figure}

When training with the BFGS method pre-training has limited effect due to the efficiency of the method to quickly reduce the value of the cost function. For less efficient methods, such as gradient descent, the difference can be quite pronounced as boundary data pre-training gives a good starting points for the iterations. There is, however, one prominent effect. The time until the first line search fail is greatly reduced by boundary data pre-training for the diffusion problem which is significantly more ill-conditioned than the advection problem. In Figure~\ref{convall} we show the convergence history of BFGS until the first line search fail with and without pre-training. That line search fails is the main bottleneck of the BFGS method and reducing them is a major gain.

\subsection{Number of hidden layers}
Most earlier works, as for example in the pioneering work by Lagaris et al. \cite{lagarisold, lagarisnew}, have been focused on neural networks with a single hidden layer. Single hidden layer networks are sufficient to approximate any continuous function to any accuracy by having a sufficient amount of neurons in the hidden layer \cite{Hornik1989359, Hornik1990551, Li1996327}. However, by fixing the capacity of the network and increasing the number of hidden layers we can see a dramatic decrease in the number of iterations required to reach a desired level of accuracy.

\begin{figure}[H]
\centering
\includegraphics[width = 0.75\textwidth]{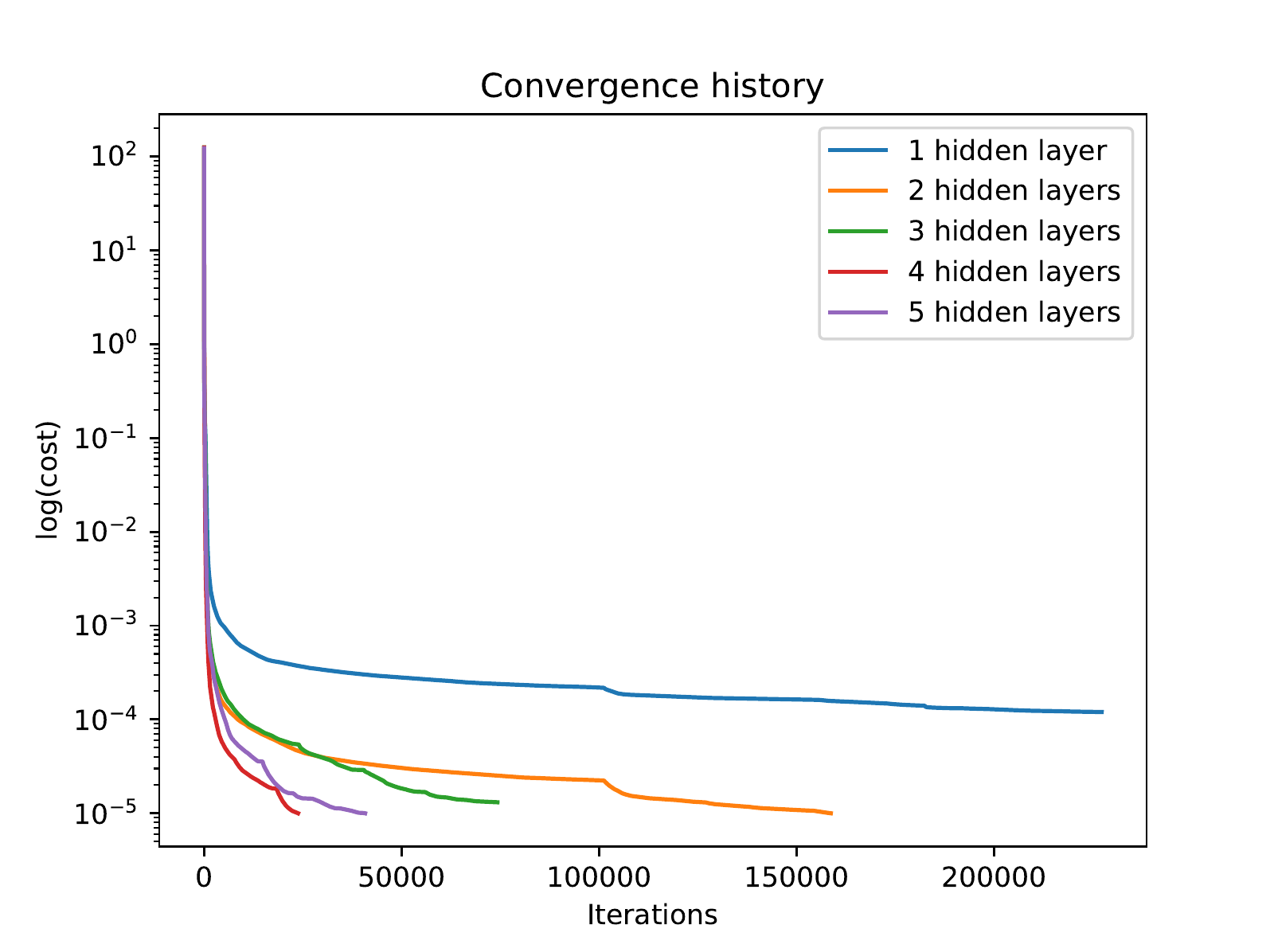}
\caption{Number of iterations required for increasing network depth.}
\label{layerconv}
\end{figure}

In the following we solve the diffusion problem in 2D as before (with boundary data pre-training) using networks with 1, 2, 3, 4 and 5 hidden layers with 120, 20, 14, 12, and 10, neurons in each hidden layer, respectively. This gives networks with 481, 501, 477, 517, and 481 trainable parameters (weights and biases), respectively, which we consider comparable. The networks are trained until the cost is less than $10^{-5}$ and we record the number of required iterations. The result can be seen in Figure~\ref{layerconv}. We can clearly see how the number of required iterations decreases as we increase the number of hidden layers. At five hidden layers we are starting to experience the vanishing gradient problem \cite{difficultdeep} and the convergence starts to deteriorate. Note that when using a  single hidden layer we did not reach the required tolerance until we exceeded the maximum number of allowed iterations.
\section{Summary and conclusion}\label{sec7}
This paper presents a method for solving advection and diffusion type PDEs in complex geometries using deep feedforward artificial neural networks. By following the derivations of the modified backpropagation algorithms outlined here, one can extend this work to include non-linear systems of PDEs of arbitrary order.

We show examples of advection and diffusion type PDEs  in 1D and 2D with one example being a highly complex polygon for which traditional mesh based methods are infeasible.

It is shown that increasing the number of hidden layers is beneficial for the number of training iterations required to reach a certain accuracy. The effects will be even more pronounced when adding deep learning techniques that prevents the vanishing gradient problem. Extending the feedforward network to more complex configurations, and to develop even deeper networks, using for example convolutional layers, dropout, and batch normalization is the topic for future works.

\section{Acknowledgements}
The authors were partially supported by a grant from the G{\"o}ran Gustafsson Foundation for Research in Natural Sciences and Medicine.

\bibliographystyle{abbrv}
\bibliography{citings}

\appendix

\section{Advection problems}
For advection problems we have a PDE of the form
\begin{equation}
Lu = \nabla \cdot (Vu) = f
\label{advecpde}
\end{equation}
for some (non-linear) matrix coefficient $V$. When the advection operator $L$ acts on $\hat{u}$ we need to compute the gradient of the ANN with respect to the input. The gradients can be computed by taking partial derivatives of (\ref{feedforward}). We get in component form
\begin{equation}
\begin{aligned}
\frac{\partial y^L_j}{\partial x_i} &= \sigma'(z^L_j) \frac{\partial z^L_j}{\partial x_i} \\
\frac{\partial z^L_j}{\partial x_i} &= \sum_k w^{L}_{jk} \sigma'_{L-1}(z^{L-1}_k) \frac{\partial z^{L-1}_k}{\partial x_i} \\
\frac{\partial z^{L-1}_j}{\partial x_i} &= \sum_k w^{L-1}_{jk} \sigma'_{L-2}(z^{L-2}_k) \frac{\partial z^{L-2}_k}{\partial x_i} \\
&\mathrel{\makebox[\widthof{=}]{\vdots}} \\
\frac{\partial z^2_j}{\partial x_i} &= \sum_k w^2_{jk} \sigma'_1(z^1_k) \frac{\partial z^1_k}{\partial x_i} \\
\frac{\partial z^1_j}{\partial x_i} &= w^1_{ji}.
\end{aligned}
\label{feedforward1}
\end{equation}
Note the slight abuse of subscript notation in order to avoid having too many subscripts. Note also that (\ref{feedforward1}) defines a new ANN with the same weights, but no biases, as the original ANN with modified activation functions. A convenient vectorial form is obtained by dropping the subscripts and re-writing (\ref{feedforward1}) as
\begin{equation}
\begin{aligned}
\frac{\partial y^L}{\partial x_i} &= \sigma'_L(z^L) \odot \frac{\partial z^L}{\partial x_i} \\
\frac{\partial z^L}{\partial x_i} &= W^L \sigma'_{L-1}(z^{L-1}) \odot \frac{\partial z^{L-1}}{\partial x_i} \\
\frac{\partial z^{L-1}}{\partial x_i} &= W^{L-1} \sigma'_{L-2}(z^{L-2}) \odot \frac{\partial z^{L-2}}{\partial x_i} \\
&\mathrel{\makebox[\widthof{=}]{\vdots}} \\
\frac{\partial z^2}{\partial x_i} &= W^2 \sigma'_1(z^1) \odot \frac{\partial z^1}{\partial x_i} \\
\frac{\partial z^1}{\partial x_i} &= W^1e_i
\end{aligned}
\label{feedforward1vec}
\end{equation}
where $e_i$ is the $i$th standard basis vector. To compute all gradients simultaneously we define for a vector $v$ and vector-valued function $f$ the diagonal and Jacobian matrices
\begin{equation}
\begin{aligned}
\Sigma(v) = \left[
	\begin{array}{cccc}
		v_1  & 0        & \cdots       & 0 \\
		0        & v_2  & \cdots       & 0 \\
		\vdots & \vdots & \ddots       & \vdots \\
		0        & 0       & \cdots        & v_{N}
	\end{array}
	\right], &&
J(f) = \left[
	\begin{array}{cccc}
		\frac{\partial f_1}{\partial x_1}  & \frac{\partial f_1}{\partial x_2}        & \cdots       & \frac{\partial f_1}{\partial x_N} \\
		\frac{\partial f_2}{\partial x_1}        & \frac{\partial f_2}{\partial x_2}  & \cdots       & \frac{\partial f_2}{\partial x_N} \\
		\vdots & \vdots & \vdots       & \vdots \\
		\frac{\partial f_{M-1}}{\partial x_1}        & \cdots & \frac{\partial f_{M-1}}{\partial x_{N-1}} & \frac{\partial f_{M-1}}{\partial x_N} \\
		\frac{\partial f_{M}}{\partial x_1}        & \cdots       & \frac{\partial f_{M}}{\partial x_{N-1}}        & \frac{\partial f_{M}}{\partial x_N}
	\end{array}
	\right].
\end{aligned}
\end{equation}
Using these matrices we can write all partial derivatives in the compact matrix form
\begin{equation}
\begin{aligned}
J(y^L) &= \Sigma(\sigma'_L(z^L)) J(z^L) \\
J(z^L) &= W^L \Sigma(\sigma'_{L-1}(z^{L-1})) J(z^{L-1}) \\
J(z^{L-1}) &= W^{L-1} \Sigma(\sigma'_{L-2}(z^{L-2})) J(z^{L-2}) \\
&\mathrel{\makebox[\widthof{=}]{\vdots}} \\
J(z^2) &= W^2 \Sigma(\sigma'_1(z^1)) J(z^1) \\
J(z^1) &= W^1 I_{N \times N}
\end{aligned}
\end{equation}
where $I_{N \times N}$ is the $N \times N$ identity matrix. All gradients of the ANN with respect to the input are contained in the rows of the Jacobian matrix $J(y^L)$ as
\begin{equation}
J(y^L) = \left[
	\begin{array}{ccc}
	\cdots & (\nabla y^L_1)^T & \cdots \\
	\cdots & \vdots & \cdots \\
	\cdots & (\nabla y^L_M)^T & \cdots
	\end{array}
\right].
\end{equation}

To solve the PDE (\ref{advecpde}) we need to modify the backpropagation algorithm (\ref{backprop}) to include the gradients of the residual cost function (\ref{rescost}) with respect to the network parameters. First we need to compute
\begin{equation}
\begin{aligned}
\frac{\partial y_m^L}{\partial w^l_{jk}}, && \frac{\partial y_m^L}{\partial b^l_j}, && m = 1, \ldots, M.
\end{aligned}
\end{equation}
This can be done componentwise by using backpropagation with the identity as the cost function. That is we let $C(y^L_m) = y^L_m$ and then (\ref{backprop}) becomes
\begin{equation}
\begin{aligned}
\delta^L_m &= \sigma'_L(z^L_m), &
\frac{\partial y^L_m}{\partial w^l_{jk}} &= y^{l-1}_k \delta^l_j, \\
\delta^l_j &= \sum_k w^{l+1}_{kj} \delta^{l+1}_k \sigma'_l(z^l_j), &
\frac{\partial y^L_m}{\partial b^l_j} &= \delta^l_j.
\end{aligned}
\label{backprop0}
\end{equation}
The $\delta$-terms in (\ref{backprop0}) can be written in vectorial form as
\begin{equation}
\begin{aligned}
\delta^L &= \sigma'_L(z^L), \\
\delta^l &= (W^{l+1})^T \delta^{l+1} \odot \sigma'_l(z^l).
\end{aligned}
\end{equation}
Secondly we need to compute
\begin{equation}
\begin{aligned}
\frac{\partial^2y^L_m}{\partial x_i\partial w^l_{jk}}, && \frac{\partial^2y^L_m}{\partial x_i\partial b^l_j}, && i = 1, \ldots, N, && m = 1, \ldots, M.
\end{aligned}
\end{equation}
This can be done by taking the partial derivative of (\ref{backprop0}) with respect to $x_i$. We get
\begin{equation}
\begin{aligned}
\frac{\delta^L_m}{\partial x_i} &= \sigma''_L(z^L_m) \frac{\partial z^L_m}{\partial x_i}, \\
\frac{\partial \delta^l_j}{\partial x_i} &= \sum_k w^{l+1}_{kj} \left( \frac{\partial\delta^{l+1}_k}{\partial x_i} \sigma'_l(z^l_j) + \delta^{l+1}_k \sigma''_l(z^l_j) \frac{\partial z^l_j}{\partial x_i} \right), \\
\frac{\partial^2 y^L_m}{\partial x_i\partial w^l_{jk}} &= \frac{\partial y^{l-1}_k}{\partial x_i} \delta^l_j + y^{l-1}_k\frac{\partial \delta^l_j}{\partial x_i}, \\
\frac{\partial^2 y^L_m}{\partial x_i\partial b^l_j} &= \frac{\partial \delta^l_j}{\partial x_i}.
\end{aligned}
\label{backprop1}
\end{equation}
The $\delta$-terms in (\ref{backprop1}) can again be written in vectorial form as
\begin{equation}
\begin{aligned}
\frac{\delta^L}{\partial x_i} &= \sigma''_L(z^L) \odot \frac{\partial z^L}{\partial x_i}, \\
\frac{\partial \delta^l}{\partial x_i} &= (W^{l+1})^T \left( \frac{\partial\delta^{l+1}}{\partial x_i} \odot \sigma'_l(z^l) + \delta^{l+1} \odot \sigma''_l(z^l) \odot \frac{\partial z^l}{\partial x_i} \right)
\end{aligned}
\end{equation}
or in matrix form as
\begin{equation}
\begin{aligned}
J(\delta^L) &= \Sigma(\sigma''_L(z^L)) J(z^L), \\
J(\delta^l) &= (W^{l+1})^T \left(\Sigma(\sigma'_l(z^l)) J(\delta^{l+1}) + \Sigma(\sigma''_l(z^l)) \Sigma(\delta^{l+1}) J(z^l) \right).
\end{aligned}
\end{equation}

The feedforward and backpropagation algorithms (\ref{feedforward}), (\ref{feedforward1}), (\ref{backprop0}), and (\ref{backprop1}), are sufficient to compute the solution to any first-order PDE, both linear and non-linear.

\begin{remark}
Note that all data from a standard feedforward and backpropagation pass is re-used in the computation of (\ref{feedforward1}) and (\ref{backprop1}). The complexity of computing (\ref{feedforward1}) and (\ref{backprop1}) is thus the same as when computing (\ref{feedforward}) and (\ref{backprop}) with some additional memory overhead to store the activations and weighted inputs.
\end{remark}

\section{Diffusion problems}
For diffusion problems we have a PDE of the form
\begin{equation}
Lu = \nabla \cdot (V \nabla u) = f.
\label{diffpde}
\end{equation}
When the diffusion operator acts on the ansatz we get second-order terms which we have not already computed. We take the partial derivative of (\ref{feedforward1}) with respect to $x_i$ to get
\begin{equation}
\begin{aligned}
\frac{\partial^2 y^L_m}{\partial x^2_i} &= \sigma''_L(z^L_m)\left( \frac{\partial z^L_m}{\partial x_i} \right)^2 + \sigma'_L(z^L_m)\frac{\partial^2 z^L_m}{\partial x^2_i} \\
\frac{\partial^2 z^L_j}{\partial x^2_i} &= \sum_k w^L_{jk} \left( \sigma''_{L-1}(z^{L-1}_k) \left( \frac{\partial z^{L-1}_k}{\partial x_i} \right)^2 + \sigma'_{L-1}(z^{L-1}_k) \frac{\partial ^2 z^{L-1}_k}{\partial x^2_i} \right) \\
\frac{\partial^2 z^{L-1}_j}{\partial x^2_i} &= \sum_k w^{L-1}_{jk} \left( \sigma''_{L-2}(z^{L-2}_k) \left( \frac{\partial z^{L-2}_k}{\partial x_i} \right)^2 + \sigma'_{L-2}(z^{L-2}_k) \frac{\partial ^2 z^{L-2}_k}{\partial x^2_i} \right) \\
&\mathrel{\makebox[\widthof{=}]{\vdots}} \\
\frac{\partial^2 z^2_j}{\partial x^2_i} &= \sum_k w^2_{jk} \left( \sigma''_1(z^1_k) \left( \frac{\partial z^1_k}{\partial x_i} \right)^2 + \sigma'_1(z^1_k) \frac{\partial ^2 z^1_k}{\partial x^2_i} \right) \\
\frac{\partial^2 z^1_j}{\partial x^2_i} &= 0.
\end{aligned}
\label{feedforward2}
\end{equation}
The vectorial form of (\ref{feedforward2}) can be written as
\begin{equation}
\begin{aligned}
\frac{\partial^2 y^L}{\partial x^2_i} &= \sigma''_L(z^L)\left( \frac{\partial z^L}{\partial x_i} \right)^2 + \sigma'_L(z^L)\frac{\partial^2 z^L}{\partial x^2_i} \\
\frac{\partial^2 z^L}{\partial x^2_i} &= W^L \left( \sigma''_{L-1}(z^{L-1}) \odot \left( \frac{\partial z^{L-1}}{\partial x_i} \right)^2 + \sigma'_{L-1}(z^{L-1}) \odot \frac{\partial ^2 z^{L-1}}{\partial x^2_i} \right) \\
\frac{\partial^2 z^{L-1}}{\partial x^2_i} &= W^{L-1} \left( \sigma''_{L-2}(z^{L-2}) \odot \left( \frac{\partial z^{L-2}}{\partial x_i} \right)^2 + \sigma'_{L-2}(z^{L-2}) \odot \frac{\partial ^2 z^{L-2}}{\partial x^2_i} \right) \\
&\mathrel{\makebox[\widthof{=}]{\vdots}} \\
\frac{\partial^2 z^2}{\partial x^2_i} &= W^2 \left( \sigma''_1(z^1) \odot \left( \frac{\partial z^1}{\partial x_i} \right)^2 + \sigma'_1(z^1) \odot \frac{\partial ^2 z^1}{\partial x^2_i} \right) \\
\frac{\partial^2 z^1}{\partial x^2_i} &= 0_{N}
\end{aligned}
\end{equation}
where $0_N = [0, \ldots, 0]^T$ denotes the zero vector with $N$ elements. To write the matrix form of (\ref{feedforward2}) we define $J^2(f)$ to be the matrix of non-mixed second partial derivatives of a vector-valued function $f$. That is
\begin{equation}
J^2(f) = \left[
	\begin{array}{cccc}
		\frac{\partial^2 f_1}{\partial x^2_1}  & \frac{\partial^2 f_1}{\partial x^2_2}        & \cdots       & \frac{\partial^2 f_1}{\partial x^2_N} \\
		\frac{\partial^2 f_2}{\partial x^2_1}        & \frac{\partial^2 f_2}{\partial x^2_2}  & \cdots       & \frac{\partial^2 f_2}{\partial x^2_N} \\
		\vdots & \vdots & \vdots       & \vdots \\
		\frac{\partial^2 f^2_{M-1}}{\partial x_1}        & \cdots & \frac{\partial^2 f_{M-1}}{\partial x^2_{N-1}} & \frac{\partial^2 f_{M-1}}{\partial x^2_N} \\
		\frac{\partial^2 f_{M}}{\partial x^2_1}        & \cdots       & \frac{\partial^2 f_{M}}{\partial x^2_{N-1}}        & \frac{\partial^2 f_{M}}{\partial x^2_N}
	\end{array}
	\right].
\end{equation}
The matrix form can then be written as
\begin{equation}
\begin{aligned}
J^2(y^L) &= \Sigma(\sigma''(z^L)) J(z^L)^2 + \Sigma(\sigma'_L(z^L)) J^2(z^L) \\
J^2(z^L) &= W^L (\Sigma(\sigma''_{L-1}(z^{L-1})) J(z^{L-1})^2 + \Sigma(\sigma'_{L-1}(z^{L-1})) J^2(z^{L-1})) \\
J^2(z^{L-1}) &= W^{L-1} (\Sigma(\sigma''_{L-2}(z^{L-2})) J(z^{L-2})^2 + \Sigma(\sigma'_{L-2}(z^{L-2})) J^2(z^{L-2})) \\
&\mathrel{\makebox[\widthof{=}]{\vdots}} \\
J^2(z^2) &= W^2 (\Sigma(\sigma''_1(z^1)) J(z^1)^2 + \Sigma(\sigma'_1(z^1)) J^2(z^1)) \\
J^2(z^1) &= W^1Z_{N \times N}
\end{aligned}
\end{equation}
where $Z_{N \times N}$ is the zero matrix of dimension $N \times N$.

The gradient of the residual cost function (\ref{rescostgrad}) requires that we compute the third-order terms
\begin{equation}
\begin{aligned}
\frac{\partial^3 y^L_m}{\partial x^2_i\partial w^l_{jk}}, && \frac{\partial^3 y^L_m}{\partial x^2_i\partial b^l_j}, && i = 1, \ldots, N, && m = i, \ldots, M.
\end{aligned}
\end{equation}
These can be computed by taking the partial derivative of (\ref{backprop1}) with respect to $x_i$ to get
\begin{equation}
\begin{aligned}
\frac{\partial^2 \delta^L_j}{\partial x^2_i} &= \sigma^{(3)}_L(z^L_j) \frac{\partial z^L_j}{\partial x_i} + \sigma''_L(z^L_j) \frac{\partial^2 z^L_j}{\partial x^2_i}, \\
\frac{\partial^2 \delta^l_j}{\partial x^2_i} &= \sum_k w^{l+1}_{jk} \left( \frac{\partial^2 \delta^{l+1}_k}{\partial x^2_i} \sigma'_l(z^l_j) + 2 \frac{\partial \delta^{l+1}_k}{\partial x_i} \sigma''_l(z^l_j) \frac{\partial z^l_j}{\partial x_i} \right) \\
&\mathrel{\makebox[\widthof{=}]{+}} \sum_k w^{l+1}_{jk} \left( \delta^{l+1}_k \sigma^{(3)}_l(z^l_j) \left( \frac{\partial z^l_j}{\partial x_i} \right)^2 + \delta^{l+1}_k \sigma''_k(z^j_l) \frac{\partial z^l_j}{\partial x_i} \right), \\
\frac{\partial^3 y^L_m}{\partial x^2_i w^l_{jk}} &= \frac{\partial^2 y^{l-1}_k}{\partial x^2_i} \delta^l_j + 2\frac{\partial y^{l-1}_k}{\partial x_i} \frac{\partial \delta^l_j}{\partial x_i} + y^{l-1}_k \frac{\partial ^2 \delta^{l-1}_k}{\partial x^2_i}, \\
\frac{\partial^3 y^L_m}{\partial x^2_i b^l_j} &= \frac{\partial^2 \delta^l_j}{\partial x^2_i}.
\end{aligned}
\label{backprop2}
\end{equation}
The vectorial form of the $\delta$-terms in (\ref{backprop2}) can be written as
\begin{equation}
\begin{aligned}
\frac{\partial^2 \delta^L}{\partial x^2_i} &= \sigma^{(3)}_L(z^L) \odot \frac{\partial z^L}{\partial x_i} + \sigma''_L(z^L) \odot \frac{\partial^2 z^L}{\partial x^2_i}, \\
\frac{\partial^2 \delta^l}{\partial x^2_i} &= (W^{l+1})^T \left( \frac{\partial^2 \delta^{l+1}}{\partial x^2_i} \odot \sigma'_l(z^l) + 2 \frac{\partial \delta^{l+1}}{\partial x_i} \odot \sigma''_l(z^l) \odot \frac{\partial z^l}{\partial x_i} \right) \\
&\mathrel{\makebox[\widthof{=}]{+}} (W^{l+1})^T\left( \delta^{l+1} \odot \sigma^{(3)}_l(z^l) \odot \left( \frac{\partial z^l}{\partial x_i} \right)^2 + \delta^{l+1} \odot \sigma''_l(z^l) \odot \frac{\partial z^l}{\partial x_i} \right)
\end{aligned}
\end{equation}
and the matrix form as
\begin{equation}
\begin{aligned}
J^2(\delta^L) &= \Sigma(\sigma^{(3)_L}(z^L)) J(z^L) + \Sigma(\sigma''_{L}(z^L)) J^2(z^L), \\
J^2(\delta^l) &= (W^{l+1})^T \left( \Sigma(\sigma'_l(z^l)) J^2(\delta^{l+1}) - 2 \Sigma(\sigma''_l(z^l)) J(\delta^{l+1}) J(z^l) \right) \\
&+ (W^{l+1})^T \left( \Sigma(\sigma^{(3)}_l(z^l)) \Sigma(\delta^{l+1}) J(z^l)^2 + \Sigma(\sigma''_l(z^l)) \Sigma(\delta^{l+1}) J(z^l) \right).
\end{aligned}
\end{equation}

The feedforward and backpropagation algorithms (\ref{feedforward}), (\ref{feedforward1}), (\ref{backprop0}), (\ref{backprop1}), (\ref{feedforward2}), and (\ref{backprop2}) can be used to solve any second-order PDE, both linear and non-linear. To compute mixed derivatives (\ref{feedforward2}) and (\ref{backprop2}) have to be modified by taking the appropriate partial derivative of (\ref{feedforward1}) and (\ref{backprop1}). Higher-order PDEs can be solved by repeated differentiation of the feedforward and backpropagation algorithms. Note that to compute second-order derivatives the weighted inputs, partial derivatives of the weighted inputs, activations, and partial derivatives of the activations needs to be re-computed or stored from the previous forward and backward passes.

\end{document}